\documentclass[a4paper,fleqn]{cas-dc}

\usepackage{times} 
\usepackage{xcolor}
\usepackage{soul}
\usepackage[utf8]{inputenc}
\usepackage[small]{caption}
\usepackage{multirow}
\usepackage{graphicx}  %Required
\usepackage{helvet}
\usepackage{courier}
\usepackage{url}
\usepackage{float}
\usepackage{color}
\usepackage{comment}
\usepackage{amsmath}
\usepackage{amssymb}
\usepackage{booktabs}
\usepackage{verbatim}
\usepackage{amsthm}
\usepackage{url}

\usepackage[loose]{subfigure}
\usepackage{epstopdf}

\usepackage{float}

\usepackage[justification=centering]{caption}  
\usepackage{booktabs} % add
\usepackage[figuresright]{rotating}
\usepackage{makecell}
\usepackage{geometry}
\usepackage{amssymb}
\usepackage{soul} 
\usepackage{color, xcolor} 
\soulregister{\cite}{7}
\soulregister{\ref}{7}

\usepackage[round,authoryear]{natbib}
\usepackage{subfigure}
\usepackage{amsmath}
\usepackage{soul}
\usepackage{color, xcolor}

\usepackage{bbm}
\usepackage{pifont}
\usepackage{algorithmicx,algorithm}
\usepackage[noend]{algpseudocode}
\usepackage{listings}

\soulregister\cite7
\soulregister\citep7
\soulregister\ref7

\def\tsc#1{\csdef{#1}{\textsc{\lowercase{#1}}\xspace}}
\tsc{WGM}
\tsc{QE}
\begin{document}

\begin{titlepage}
\begin{center}
\vspace*{1cm}

\textbf{ \large BiPC: Bidirectional Probability Calibration for Unsupervised Domain Adaption}

\vspace{1.5cm}

% Author names and affiliations
Wenlve Zhou$^{a}$(eewenlvezhou@mail.scut.edu.cn), Zhiheng Zhou$^{a}$(zhouzh@scut.edu.cn), Junyuan Shang$^{a}$(eeshangjy@mail.scut.edu.cn), Chang Niu$^{a}$(eeniu@mail.scut.edu.cn), Mingyue Zhang$^{b}$(2023001040@usc.edu.cn), Xiyuan Tao$^{a}$(eetaoxy@mail.scut.edu.cn), Tianlei Wang$^{c}$(tianlei.wang@aliyun.com)\\

\hspace{10pt}

\begin{flushleft}
\small  
$^{a}$ School of Electronic and Information Engineering, South China University of Technology, Guangzhou 510640, China \\
$^{b}$ School of Computing, University of South China, Hengyang, 421001, China \\
$^{c}$ School of Electronics and Information Engineering, Wuyi University, Jiangmen 529020, China
%
%\begin{comment}
%Clearly indicate who will handle correspondence at all stages of refereeing and publication, also post-publication. Ensure that phone numbers (with country and area code) are provided in addition to the e-mail address and the complete postal address. Contact details must be kept up to date by the corresponding author.
%\end{comment}

\vspace{1cm}
\textbf{Corresponding Author:} \\
Zhiheng Zhou \\
School of Electronic and Information Engineering, South China University of Technology, Guangzhou 510640, China \\
Tel: 13660788980(+86 and 020) \\
Email: zhouzh@scut.edu.cn

\end{flushleft}        
\end{center}
\end{titlepage}

\let\WriteBookmarks\relax
\def\floatpagepagefraction{1}
\def\textpagefraction{.001}

\shorttitle{W. Zhou et~al./ Expert Systems with Applications}
\shortauthors{zwl et~al.}

\title [mode = title]{BiPC: Bidirectional Probability Calibration for Unsupervised Domain Adaption}

\author[a]{Wenlve Zhou}[orcid=0000-0002-7500-5581]
\ead{eewenlvezhou@mail.scut.edu.cn}

\author[a]{Zhiheng Zhou}[orcid=0000-0003-4040-0175] 
\cormark[1]
\ead{zhouzh@scut.edu.cn}

\author[a]{Junyuan Shang}[orcid=0000-0003-4301-750X]
\ead{eeshangjy@mail.scut.edu.cn}

\author[a]{Chang Niu}[orcid=0000-0002-7426-1479]
\ead{eeniu@mail.scut.edu.cn}

\author[b]{Mingyue Zhang}[orcid=0000-0002-2783-9303]
\ead{2023001040@usc.edu.cn}

\author[a]{Xiyuan Tao}[orcid=0000-0001-5075-0545]
\ead{eetaoxy@mail.scut.edu.cn}

\author[c]{Tianlei Wang}[orcid=0000-0002-6983-0788]
\ead{tianlei.wang@aliyun.com}

\address[a]{School of Electronic and Information Engineering, South China University of Technology, Guangzhou 510640, China}
\address[b]{School of Computing, University of South China, Hengyang, 421001, China}
\address[c]{School of Electronics and Information Engineering, Wuyi University, Jiangmen 529020, China}

\cortext[0]{\textbf{The definitive version of this paper can be found at: 10.1016/j.eswa.2024.125460}}
\cortext[1]{Corresponding author.}

\begin{abstract}
Unsupervised Domain Adaptation (UDA) leverages a labeled source domain to solve tasks in an unlabeled target domain. While Transformer-based methods have shown promise in UDA, their application is limited to plain Transformers, excluding Convolutional Neural Networks (CNNs) and hierarchical Transformers. To address this issues, we propose Bidirectional Probability Calibration (BiPC) from a probability space perspective. We demonstrate that the probability outputs from a pre-trained head, after extensive pre-training, are robust against domain gaps and can adjust the probability distribution of the task head. Moreover, the task head can enhance the pre-trained head during adaptation training, improving model performance through bidirectional complementation. Technically, we introduce Calibrated Probability Alignment (CPA) to adjust the pre-trained head's probabilities, such as those from an ImageNet-1k pre-trained classifier. Additionally, we design a Calibrated Gini Impurity (CGI) loss to refine the task head, with calibrated coefficients learned from the pre-trained classifier. BiPC is a simple yet effective method applicable to various networks, including CNNs and Transformers. Experimental results demonstrate its remarkable performance across multiple UDA tasks. Our code will be available at: https://github.com/Wenlve-Zhou/BiPC.
\end{abstract}

\maketitle

\section{Introduction}
In recent years, deep learning has demonstrated impressive power in a wide range of tasks (\cite{ref1, ref2, ref3}), yet it has not performed as well when applied to the actual target dataset as neural networks are sensitive to domain gaps. To address this issue, unsupervised domain adaptation (UDA) (\cite{ref34}) has been introduced to transfer knowledge from a labeled source domain to an unlabeled target domain. Currently, popular UDA methods can be categorized into feature alignment and Transformer-based approaches. These techniques have significantly advanced the field.

\begin{figure}[!t]
	\setlength{\abovecaptionskip}{0.cm}
	\setlength{\belowcaptionskip}{-0.cm}
	\centering
	\includegraphics[width=2.6in]{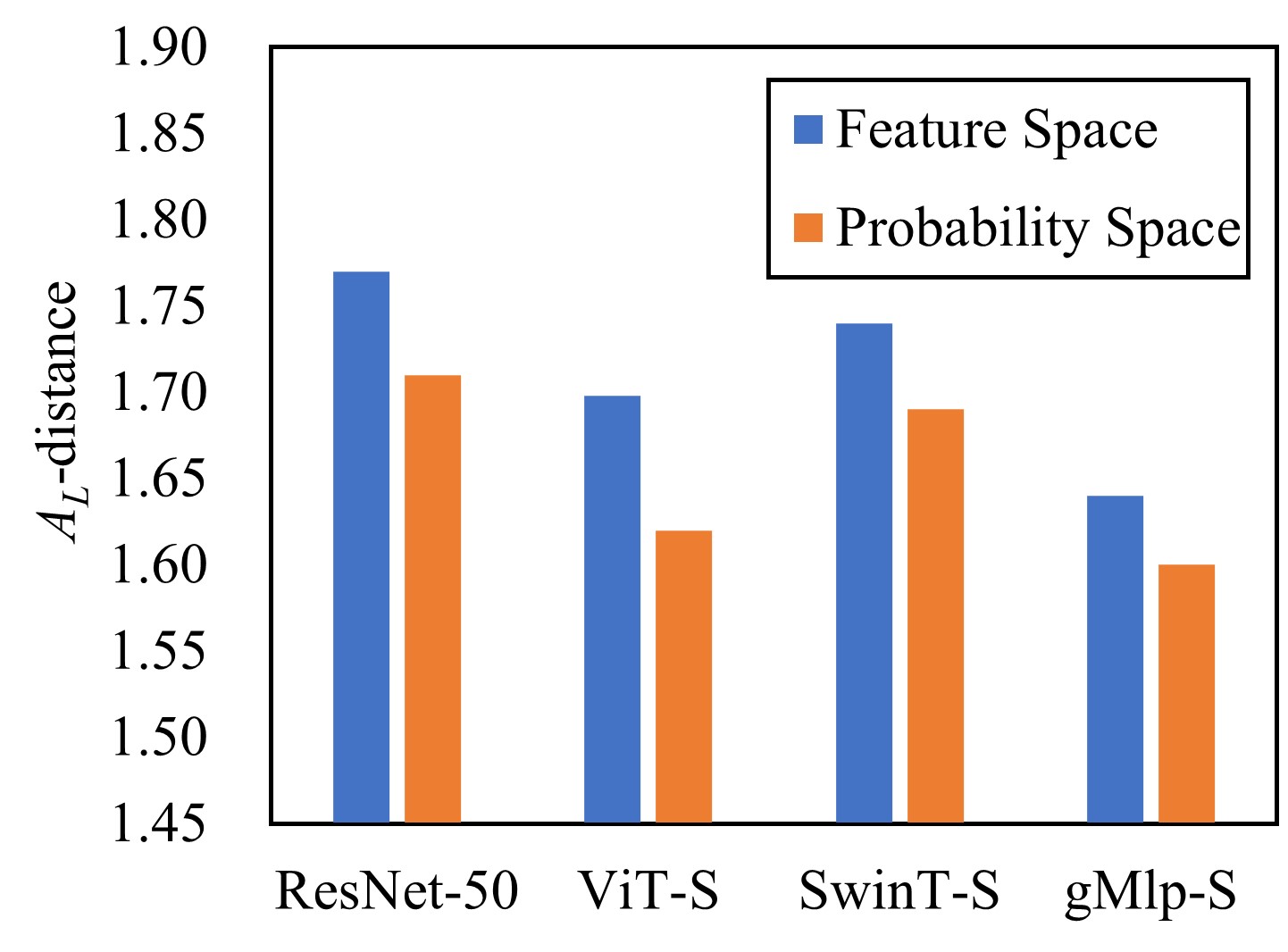}
	\caption{$A_L$-distance on feature space and probability space of different architectures between Art and Clipart from Office-Home based on ImageNet pre-trained models. The feature space refers to the distribution before the pre-trained classifier and the probability space represents its output. It can be seen that the probability space of the pre-trained model has a smaller domain gap.}
	\label{fig1}
\end{figure}

Feature alignment techniques focus on aligning the feature space of the source and target domains, aiming to learn domain-invariant representation (\cite{ref4, ref5, ref6, ref7}). Popular paradigms include Maximum Mean Discrepancy (MMD) (\cite{ref4}) and adversarial-based methods (\cite{ref5}). The former aligns feature in Hilbert space using kernel methods, while the latter learns invariant feature through adversarial learning with a discriminator. However, these methods do not consider distribution matching between categories. To address the challenge of global alignment, researchers have turned their attention to category alignment, which aims to align representation of the same class between the source and target domains (\cite{ref8, ref9}). A common practice is to estimate target domain labels using pseudo annotation (\cite{ref22}), combined with MMD for category-invariant feature learning. These feature alignment methods are predominantly based on Convolutional Neural Networks (CNNs) and have shown promising results.

However, category alignment methods are not robust against noisy pseudo-labels due to the inductive bias of CNNs. With the successful adoption of Transformers in Natural Language Processing (NLP) (\cite{ref10, ref11}) and Computer Vision (CV) (\cite{ref12, ref13}), recent studies have discovered that the global attention mechanism in Transformers is resilient to noise in pseudo-labels generated for the target domain. Furthermore, Transformers have been found to possess significant potential for UDA tasks, surpassing feature alignment methods (\cite{ref14, ref15, ref16}). Nevertheless, UDA techniques aim to empower the system with adaptation, rather than being restricted to a fixed network structure. To the best of our knowledge, Transformer-based algorithms like CDTrans (\cite{ref14}) cannot be directly applied to CNNs, nor can they be generalized to hierarchical Transformers. Therefore, we need to explore whether there exists a more universal technique to enhance domain adaptation capabilities for different networks, be it CNNs or Transformers. One straightforward idea is to directly apply feature alignment to Transformers. However, Yang et al. (\cite{ref15}) indicated that the differences in the inductive biases of the backbones lead to variations in the feature space. Hence, the feature alignment methods designed for CNNs have shown insufficient improvements for Transformers.

\begin{figure*}[!t]
	\centering
	\includegraphics[width=6in]{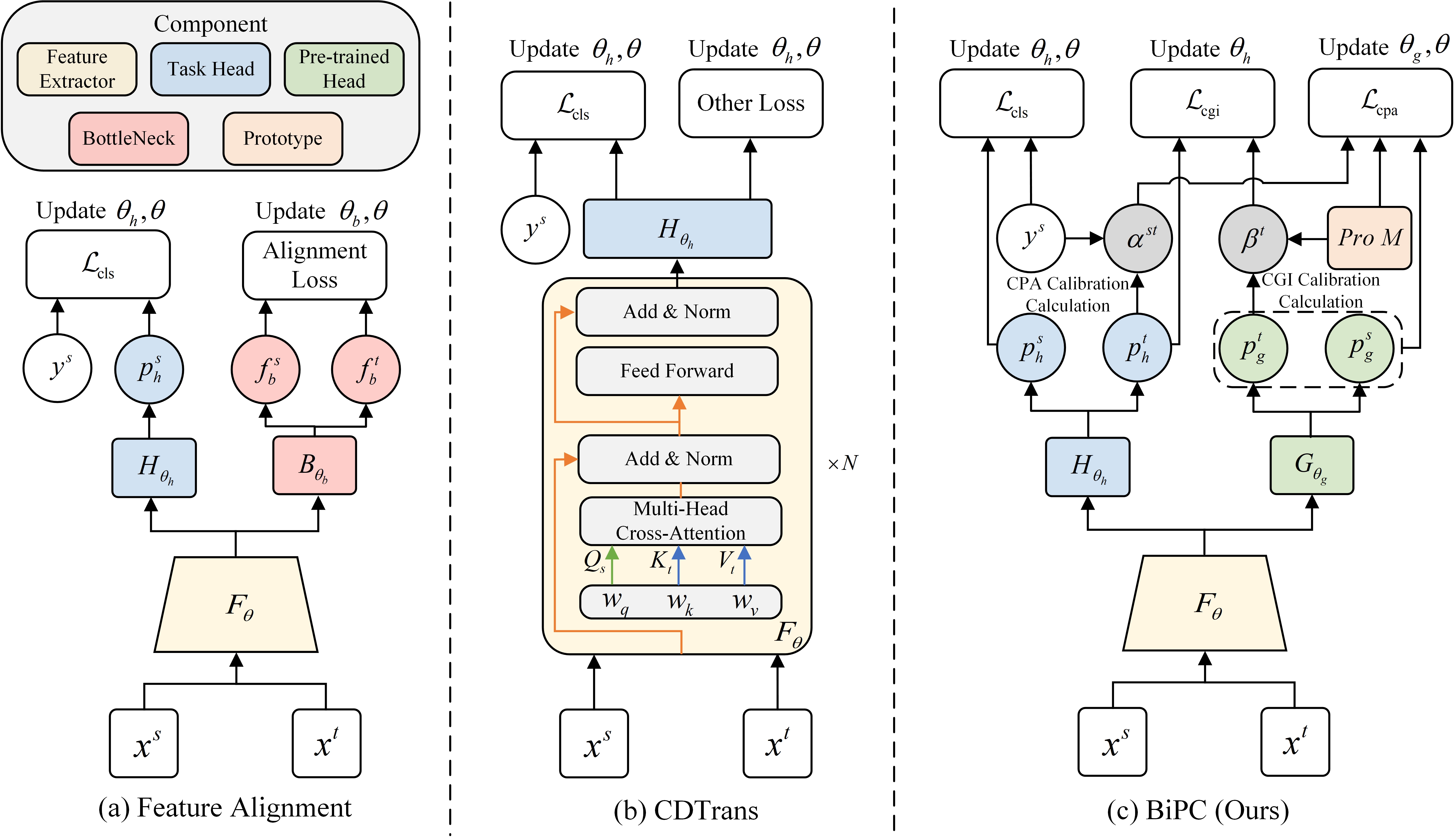}
	\caption{Pipeline of the previous methods and BiPC. (a) This paradigm typically involves compressing the feature space using a bottleneck layer to align the source and target domains through an alignment loss function, such as MMD (\cite{ref4}). (b) Transformer-based UDA. CDTrans (\cite{ref14}) serves as an example of this paradigm, which achieves invariant feature learning through a specifically designed cross-domain attention mechanism. However, it is only applicable to plain Transformers, like ViT (\cite{ref12}). (c) The proposed BiPC. This approach combines effectiveness and flexibility and can be adapted to different architectures, as described in Section 3.}
	\label{fig2}
\end{figure*}

Considering these challenges, we shift our focus towards the probability space derived from task-specific top layers, such as the ImageNet pre-trained classifier (\cite{ref17}). \textbf{We propose that probability space can offer more concise representation to guide adaptation training, and this phenomenon holds true for different architectures.} The rationale behind this is that the pre-trained classifier encompasses a significant number of parameters, enabling it to better capture intra-class and inter-class relationships (\cite{ref18}). To validate our hypothesis, we compute the domain distance between the Art and Clipart subsets from the Office-Home dataset (\cite{ref21}) in both the feature space and the probability space using the ImageNet pre-trained model. For the experiments, we investigate ResNet-50 (\cite{ref19}), ViT-S (\cite{ref12}), SwinT-S (\cite{ref13}), and gMlp-S (\cite{ref20}), all pre-trained on ImageNet-1K, while employing the $A_L$-distance metric proposed in (\cite{ref9}) to measure the domain gap. Figure 1 illustrates that, even across different pre-trained models, the probability space presents a more effective representation of sample relationships, exhibiting a smaller domain gap compared to the feature space.

Hence, based on the previous analysis, we propose a simple but effective UDA method called Bidirectional Probability Calibration (BiPC) to address the limitations of existing approaches. For pre-trained head training, we introduce a Calibrated Probability Alignment (CPA) to calibrate the probability space, i.e., the distribution of the pre-trained head output. The calibration coefficient is learned from the target pseudo-label and the ground truth of source domain. Simultaneously, the relationship between samples in the probability space offers valuable information for adaptation training in the task head. To improve the distribution of the task head, we introduce a Calibrated Gini Impurity (CGI) where the calibration coefficient is learned from the probability space. Experiments show that BiPC can obtain remarkable results on popular domain adaption benchmarks.

Our main contribution can be summarized as follows:

(1) We propose a simple but effective UDA method based on the probability space, which can mitigate model performance degradation due to domain gap.

(2) A calibrated probability alignment loss is introduced which aligns the probability distribution via calibrated coefficient calculated from the source label and the target pseudo-label for pre-trained head tuning.

(3) To calibrate the distribution of task head, we design a calibrated Gini impurity loss for pseudo-label learning where the calibrated coefficient is learned from the relationship of probability space from pre-trained head.

(4) Extensive experiments show that BiPC provides a significant improvement on various backbones and achieves compelling performance in comparison with the feature alignment and Transformer-based counterparts. Our approach can even be applied to partial-set domain adaptation (PDA), achieving state-of-the-art (SoTA) on both UDA and PDA tasks.

The rest of this article is organized as follows: After a brief introduction to related work in Section 2, the proposed architecture is elaborated in Section 3. In Section 4, the experimental results are displayed to discuss and verify the effectiveness of our method, followed by the complete conclusion in Section 5. The main notations of this paper are summarized in Table 1.

\section{Related Work}
In this section, we provide a concise survey of previous methods in unsupervised domain adaptation across three key aspects: Feature Alignment UDA, Transformer-based UDA, and Pseudo Labeling. The first part introduces global alignment and category alignment methods. As Transformer are a relatively new technology for domain adaptation, their mechanisms, along with relevant UDA methods, are briefly described in the second part. Additionally, the article's presented CGI relates to pseudo-labeling, which is introduced in the third part.

\subsection{Feature Alignment UDA}
Since neural networks are sensitive to domain gap, unsupervised domain adaptation is the research of great interest that has been successfully applied in various areas such as ECG classification (\cite{ref23}), medical segmentation (\cite{ref24}), and fault diagnosis (\cite{ref8}).

In previous studies, CNNs have shown the potential to learn invariant and transferable feature (\cite{ref26,ref27}). Therefore, feature alignment methods (as depicted in Figure 2(a)) were a popular paradigm formerly. MMD (\cite{ref4,ref28}) is the classical technique, which deploys with linear kernel or multi-kernel to reduce the domain shift in the feature space. Central Moment Discrepancy (CMD) is a similar metric that Zellinger et al. (\cite{ref29}) proposed and proved the convergence on compact intervals. Not satisfied with first-order statistics, Sun et al. (\cite{ref6}) proposed deep coral matching based on second-order statistics. Besides, adversarial learning (\cite{ref30}) can also be applied to match two distributions. Ganin et al. (\cite{ref5}) are the pioneers that introduce a discriminator to measure the discrepancy of the domain gap. They take the backbone as the counterpart of the discriminator and update them by adversarial loss (\cite{ref30}). To accelerate the convergence of the model, Chadha et al. (\cite{ref31}) transfer various tricks from the GAN such as multi-task discriminator, feature disentanglement.

% Table generated by Excel2LaTeX from sheet 'Table1'
\begin{table}[htbp]
	\scriptsize
	\setlength{\abovecaptionskip}{0cm}  %段前
	\setlength{\belowcaptionskip}{-0.0cm} %段后
	\centering
	\caption{Main Notations of the Method}
	\begin{tabular}{ll}
		\toprule
		\textbf{Symbol} & \textbf{Explanation} \\
		\midrule
		$x^{s}, y^{s}$ & Source data and related label \\
		\midrule
		$x^{t}, \tilde{y}$ & Target data and related pseudo-label \\
		\midrule
		$F_{\theta}, G_{\theta_{g}}, H_{\theta_{h}}$ & Feature extractor, pre-trained head and task head \\
		\midrule
		$c_{1}, c_{2}$ & Classes of UDA and pre-training task \\
		\midrule
		$p_{h}^{s}, p_{h}^{t}$ & Probability of source and target data on task head \\
		\midrule
		$p_{g}^{s}, p_{g}^{t}$ & Probability of source and target data on pre-trained head \\
		\midrule
		$M$ & Prototype of source data on the pre-trained head \\
		\midrule
		$\tilde{p}_{h}^{t}$ & Transformation probability of target data on UDA task \\
		\midrule
		$\alpha^{s t}$ & Calibration coefficient of CPA\\
		\midrule
		$\beta^{t}$ & Calibration coefficient of CGI\\
		\midrule
		$\mathbb{E}_{*}$ & Expectations for the related variable \\
		\bottomrule
	\end{tabular}%
	\label{table1}%
\end{table}%

However, the aforementioned methods do not consider category matching, limiting the performance of UDA methods. Pei et al. (\cite{ref32}) capture multimode structures to enable fine-grained alignment of different data distributions based on multiple domain discriminators. Xie et al. (\cite{ref33}) present a moving semantic transfer network, learning semantic representations for unlabeled target data. The framework aligns the labeled source centroid and pseudo-labeled target centroid to achieve category matching. For more efficient training, Zhu et al. (\cite{ref9}) propose a local maximum mean discrepancy to align the relevant subdomain distributions of domain-specific layer activations.

Feature alignment methods are extremely flexible and can be embedded in different neural network architectures. However, the current approaches are designed for convolutional neural networks and have limited improvement for new architecture such as Transformers.

\subsection{Transformer-based UDA}
The Transformer, proposed in (\cite{ref10}), is designed to model sequential data in the field of NLP. With its impressive contextual modeling capabilities, it has revolutionized various domains in computer vision (\cite{ref35}). 

To bridge the gap between CV and NLP, the image will be divided into several patches and tokenized using learnable parameters. These patches will then be processed with a series of visual Transformer blocks (\cite{ref12}). Transformer blocks consist of several components, including self-attention, feed-forward network, residual connection, etc., with the self-attention mechanism being the core. In the self-attention layer, the input vector is first transformed into three different vectors with dimension $d_k$: the query vector $q$, the key vector $k$ and the value vector $v$. Vectors derived from different inputs are then packed together into three different matrices, namely, $Q$, $K$ and $V$. Subsequently, the attention function between different input vectors is calculated as follows:
\begin{eqnarray}
	\operatorname{Attention}({Q}, {K}, {V})=\operatorname{softmax}\left(\frac{{Q} {K}^{\top}}{\sqrt{d_{k}}}\right) {V}.
\end{eqnarray}

Concerned with its simplicity and powerful inductive bias, TVT introduces the Transformer architecture into UDA, highlighting the limitations of CNN-based adaptation strategies in utilizing Transformers' inductive bias, such as the attention mechanism and sequential image representation (\cite{ref15}). To overcome this challenge, TVT incorporates learned transferability into attention blocks, ensuring that the Transformer focuses on both transferable and discriminative features. Further exploring the attention mechanism, CDTrans combines cross-attention with pseudo-labels for cross-domain relationship learning, which significantly outperforms CNN-based approaches (\cite{ref14}). SSRT enhances performance by blending intermediate layer feature through consistent learning (\cite{ref16}). 

Previous Transformer-based methods (as shown in Figure 2(b)) have delved deeply into the capabilities of the architecture but have also limited flexibility. It is worth noting that these methods cannot be easily applied to CNNs or even generalized to other Transformer structures. Therefore, in this manuscript, we propose BiPC, a method that strikes a balance between effectiveness and flexibility.

\subsection{Pseudo Labeling}
Pseudo-labeling, initially proposed in semi-supervised learning (\cite{ref37}), has been introduced to domain adaptation tasks due to their similar task settings. This technique utilizes the predicted categories of unlabeled data and labeled data as supervised signals to train the model (\cite{ref22}).

Most methods focus on improving the quality of pseudo-labels as they are often affected by significant noise. Zhang et al. (\cite{ref38}) propose an adaptive threshold adjustment mechanism to select informative unlabeled data and their corresponding pseudo-labels. Kim et al. (\cite{ref39}) generate pseudo-labels using both the source domain and weakly-augmented target domain, which are then used to train the model between the source and strongly-augmented target domains. Wang et al. (\cite{ref40}) propose an uncertainty-aware pseudo-label assignment strategy without the need for a predefined threshold in order to reduce label noise.

In contrast to previous methods, CGI improve the task head with pseudo-labels and probability distribution from pre-trained head. Additionally, the generated pseudo-labels and CPA further refine the probability distribution from pre-trained classifier, resulting in improved model performance through bidirectional mutual complementation.

\section{Methods}
In the unsupervised domain adaption, data from two different distributions ${{p}_{s}}$ and ${{p}_{t}}$ will be sampled to form the source labeled domain ${{D}_{s}}=\!\!\{\!\!\text{ (}x_{i}^{s}\text{,}y_{i}^{s}\text{) }\!\!\}\!\!\text{ }_{i=1}^{{{n}_{s}}}$ and unlabeled target domain ${{D}_{t}}=\{x_{j}^{t}\}_{j=1}^{{{n}_{t}}}$ datasets, where ${{n}_{s}}$ and ${{n}_{t}}$ denote the quantity of data in each domain, and ${{y}^{s}}\in {{\mathbb{R}}^{{{n}_{s}}\times {{c}_{1}}}}$, where ${{c}_{1}}$ is the number of classes in UDA datasets. In previous methods, the pre-trained head ${G}_{{{\theta }_{g}}}(\cdot)$ is removed while the feature extractor ${F}_{\theta }(\cdot)$ is retained, picking up a novel task head ${H}_{{{\theta }_{h}}}(\cdot)$ to form the model to be trained. For convenience, both the pre-trained and task heads are defined to include the softmax function in the manuscript. The objective of UDA is to learn a domain-invariant network that adapts to the target domain using source data and labels, while having access to the data from the target domain during training.

Building upon the earlier analyses, we propose an alignment approach grounded in the probability space of the pre-trained head, denoted as the calibrated probability alignment ${{\mathcal{L}}_{\text{cpa}}}$ as introduced in Section 3.1. As depicted in Figure 2(c), this method aims to reconcile the distributions between the source and target domains, thereby refining the probability distribution of the pre-trained head. As the pre-training head remains insensitive to the domain gap (refer to Figure 1), the calibrated Gini impurity ${{\mathcal{L}}_{\text{cgi}}}$ adjusts the logit of the task head by leveraging the probability distribution derived from the pre-training head (outlined in Section 3.2). Enhancements in the model's performance within the target domain are achieved through the bidirectional calibration of both distributions.

The objective is formulated as
\begin{eqnarray}
	{{\mathcal{L}}_{\text{total}}}={{\lambda }_{1}} {{\mathcal{L}}_{\text{cls}}}+{{\lambda }_{2}} {{\mathcal{L}}_{\text{cpa}}}+{{\lambda }_{3}} {{\mathcal{L}}_{\text{cgi}}}.
\end{eqnarray}
where ${{\mathcal{L}}_{\text{cls}}}$ is the classification loss to measure the discrepancy between source logits of task head $p_{h}^{s}={{H}_{{{\theta }_{h}}}}({{F}_{\theta }}({{x}^{s}}))$, $p_{h}^{s}\in {{\mathbb{R}}^{{{n}_{s}}\times {{c}_{1}}}}$ and the source label ${{y}^{s}}$. The trade-off parameters ${{\lambda }_{1}}$, ${{\lambda }_{2}}$ and ${{\lambda }_{3}}$ are introduced to balance the multi-loss.

\subsection{Calibrated Probability Alignment}
Akin to category alignment techniques (\cite{ref32, ref33, ref34}), CPA focuses on aligning the source and target domains at the category level to refine the distribution of pre-trained head, but this alignment is performed in the probability space rather than the feature space. In order to quantify the dissimilarity between the corresponding subdomains, CPA is defined as follows:
\begin{eqnarray}
	{{\mathcal{L}}_{\text{cpa}}}\overset{\Delta }{\mathop{=}}\,{{\mathbb{E}}_{c}}[d({{\mathbb{E}}_{p_{s}^{c}}}[p_{g}^{s}],{{\mathbb{E}}_{p_{t}^{c}}}[p_{g}^{t}])].
\end{eqnarray}
where $p_{g}^{s}={{G}_{{{\theta }_{g}}}}({{F}_{\theta }}({{x}^{s}}))$ and $p_{g}^{t}={{G}_{{{\theta }_{g}}}}({{F}_{\theta }}({{x}^{t}}))$, $p_{g}^{s}\in {{\mathbb{R}}^{{{n}_{s}}\times {{c}_{2}}}}$ and $p_{g}^{t}\in {{\mathbb{R}}^{{{n}_{t}}\times {{c}_{2}}}}$, represent the probability distribution in pre-trained datasets (e. g., ImageNet-1k and ${{c}_{2}}$ denotes the number of classes, i.e. 1000). On the other hand, ${{x}^{s}}$ and ${{x}^{t}}$ refer to the instances in ${{D}_{s}}$ and ${{D}_{t}}$ respectively. Furthermore, $p_{s}^{c}$ and $p_{t}^{c}$ denote the distributions of $D_{s}^{\left( c \right)}$ and $D_{t}^{\left( c \right)}$. Finally, the discrepancy is measured using the distance metric $d$. 

Inspired by MMD (\cite{ref4}), CPA aims to minimize the gap by computing the expectation of the probability distribution in the category style. The unbiased estimator of Eq. (3) is then formulated as follows:
\begin{eqnarray}
	{{\mathcal{L}}_{\text{cpa}}}=\sum\nolimits_{i=1}^{{{n}_{s}}}{\sum\nolimits_{j=1}^{{{n}_{t}}}{\alpha _{ij}^{st} d(p_{gi}^{s},p_{gj}^{t})}}.
\end{eqnarray}
where $p_{gi}^{s}$ and $p_{gj}^{t}$ represent the instance of $p_{g}^{s}$ and $p_{g}^{t}$ respectively. The probability calibration coefficient of $x_{i}^{s}$ and $x_{j}^{t}$ is denoted by $\alpha _{ij}^{st}$, with ${{\alpha }^{st}}\in {{\mathbb{R}}^{{{n}_{s}}\times {{n}_{t}}}}$. In accordance with Eq. (4), it can be observed that the category learning paradigm is eliminated, as the calculation is implicitly incorporated into the coefficient. The computation of $\alpha _{ij}^{st}$ is performed as follows:
\begin{eqnarray}
	\alpha _{ij}^{st}=\alpha _{i}^{s} {{(\alpha _{j}^{t})}^{\top}}.
\end{eqnarray}
where $a_{i}^{s}\in {{\mathbb{R}}^{1\times {{c}_{\text{1}}}}}$ and $a_{j}^{t}\in {{\mathbb{R}}^{1\times {{c}_{\text{1}}}}}$ are denoted as the weights of $x_{i}^{s}$ and $x_{j}^{t}$ belonging to relevant class in the UDA task. Inspired by LMMD (\cite{ref9}), the weight of samples in each domain is calculated as follows:
\begin{eqnarray}
	\alpha _{i}^{s}=\frac{y_{i}^{s}}{\sum\nolimits_{k=1}^{{{n}_{s}}}{y_{k}^{s}}},
\end{eqnarray}
\begin{eqnarray}
	\alpha_{j}^{t}=\tilde{y}_{j} \frac{p_{h j}^{t}}{\sum_{k=1}^{n_{t}} p_{h k}^{t}}.
\end{eqnarray}
where $y_{i}^{s}\in {{\mathbb{R}}^{1\times {{c}_{\text{1}}}}}$ represents the one-hot vector of the ground truth for instance $x_{i}^{s}$. Regarding the samples in the target domain, $p_{hj}^{t}{=}{{H}_{{{\theta }_{h}}}}({{F}_{\theta }}(x_{j}^{t}))$ denotes the probability of task head, where $p_{hj}^{t}\in {{\mathbb{R}}^{\text{1}\times {{c}_{1}}}}$. The pseudo-label $\tilde{y}_{j}\,\in {{\mathbb{R}}^{1\times {{c}_{\text{1}}}}}$ is obtained by performing the argmax operation on $p_{hj}^{t}$ and converting it into a one-hot vector.

In addition to the correction factor, designing an appropriate distance metric $d$ is crucial for the success of CPA. A straightforward approach is to utilize the Kullback-Leibler (KL) divergence for aligning probability distributions. However, unlike the classification task where the probability distribution matches the label distribution, in CPA, two probability distributions need to be fitted to each other. Asymmetric optimization can result in unstable convergence and hinder performance. Therefore, it is worth considering whether the KL divergence can be replaced by the Jensen-Shannon (JS) divergence. The JS divergence is formulated as follows:
\begin{eqnarray}
	\begin{small}
		\begin{aligned}
			\mathrm{J} S\left(p_{g i}^{s} \| p_{g j}^{t}\right)= & 0.5 \left(p_{g i}^{s} \log \left(p_{g i}^{s} /\left(p_{g i}^{s}+p_{g j}^{t}\right)\right)\right. \\
			+ &\left.p_{g j}^{t} \log \left(p_{g j}^{t} /\left(p_{g i}^{s}+p_{g j}^{t}\right)\right)\right)+\log 2 \\
			= & 0.5 \left(p_{g i}^{s}  \log \left(p_{g i}^{s}\right)-p_{g i}^{s} \log \left(p_{g i}^{s}+p_{g j}^{t}\right)\right. \\
			+ & \left.p_{g j}^{t} \log \left(p_{g j}^{t}\right)-p_{g j}^{t} \log \left(p_{g i}^{s}+p_{g j}^{t}\right)\right)+\log 2
		\end{aligned}.
	\end{small}
\end{eqnarray}

Yet when minimizing the entropy term $p_{gi}^{s} \log (p_{gi}^{s})$ and $p_{gj}^{t} \log (p_{gj}^{t})$ in Eq. (8), the probability distribution tends to be sharp, destroying the relevant information in the probability space. Consequently, to address this issue, the entropy and constant terms are removed, and the distance metric $d$ is defined as follows:
\begin{eqnarray}
	\begin{aligned}
		\scalebox{0.93}{
			$d(p_{gi}^{s},p_{gj}^{t})=-0.5(p_{gi}^{s} \log (p_{gi}^{s}+p_{gj}^{t})+p_{gj}^{t} \log (p_{gi}^{s}+p_{gj}^{t})).$
		}
	\end{aligned}
\end{eqnarray}

Moreover, as the optimization of Eq. (9) involves cross-domain alignment, there is a risk of falling into shortcut learning, which essentially means that the optimization process may become overly reliant on simply averaging the class-wise probabilities of the source and target domains. This can disrupt the distribution of data in the probability space. To prevent this, a regularization term is introduced:
\begin{eqnarray}
	\begin{aligned}
		r(p_{g}^{s},M)=\sum\nolimits_{c=1}^{{{c}_{1}}}{\sum\nolimits_{i=1}^{{{n}_{s}}}{\mathbbm{1}_{[y_{i}^{s}=c]} {{M}_{c}}}}\log \frac{{{M}_{c}}}{p_{gi}^{s}}.
	\end{aligned}
\end{eqnarray}
where $\mathbbm{1}_{[\cdot]}$ is the indicator function, and $M\in {{\mathbb{R}}^{{{c}_{1}}\times {{c}_{2}}}}$ represents the prototype of the source domain data in the probability space. Intuitively, each row of $M$ represents the center of probability for data with the same label in the UDA task, within the output space of the pre-trained head. Specifically, we partition the training and validation sets of the source domain datasets in a 1:1 ratio and utilize the category relationship learning, Algorithm 1 in (\cite{ref18}) to obtain the prototype $M$. In Eq. (10), the optimization focuses on minimizing the KL divergence between the probability distribution of the source domain and the prototype. This ensures that the data aligns closely with the prototype, thereby avoiding shortcut learning. Finally, Eq. (4) can be rewritten as follows:
\begin{eqnarray}
	\begin{aligned}
		{{\mathcal{L}}_{\text{cpa}}}\text{=}\sum\nolimits_{i=1}^{{{n}_{s}}}{\sum\nolimits_{j=1}^{{{n}_{t}}}{\alpha _{ij}^{st} d(p_{gi}^{s},p_{gj}^{t})}}+r(p_{g}^{s},M).
	\end{aligned}
\end{eqnarray}

\subsection{Calibrated Gini Impurity}
The probability space derived from the pre-training dataset can help minimize the domain gap by incorporating abundant intra-class and inter-class information (\cite{ref18}), as illustrated in Figure 1. We intend to integrate the probability distribution from the pre-trained head with the pseudo-labeling technique specifically for the task of training the head portion. A common pseudo-labeling technique is to minimize the uncertainty of predictions on unlabeled data using Gibbs entropy. However, it has been shown that Gibbs entropy can lead to overconfident predictions (\cite{ref41}). Therefore, we employ a weaker penalization method called Gini impurity (GI) as the foundation for CGI. The formulation of CGI is as follows:
\begin{eqnarray}
	\begin{aligned}
		\text{GI} (p_{h}^{t})=\sum\nolimits_{j=1}^{{{n}_{t}}}{1-\sum\nolimits_{c=1}^{{{c}_{1}}}{{{(p_{hjc}^{t})}^{2}}}}.
	\end{aligned}
\end{eqnarray}
where $p_{hjc}^{t}$ is the probability of class $c$ in the task head.

Although Gini impurity helps to prevent overconfidence in predictions on unlabeled data, the results are still heavily influenced by the task head learned from the labeled source data, leading to potential bias. Considering that the pre-trained head ${G}_{{{\theta }_{g}}}(\cdot)$ is trained on large-scale pre-training datasets and can effectively capture the relationship across inter-domain data, we aim to correct the Gini impurity using the distribution of the probability space. Since the probability space alone cannot directly describe the relationship between samples, it needs to be transformed using the prototype $M$ and the target domain distribution $p_{g}^{t}$ from the pre-trained head. This transformation is expressed as follows:
\begin{eqnarray}
	\begin{aligned}
		\tilde{p}_{h j c}^{t}=\frac{\exp \left(-\mathrm{KL}\left(M_{c} \| p_{g j}^{t}\right)\right)}{\sum_{k=1}^{c_1} \exp \left(-\mathrm{KL}\left(M_{k} \| p_{g j}^{t}\right)\right)}.
	\end{aligned}
\end{eqnarray}
where $\tilde{p}_{h j c}^{t} \in \mathbb{R}^{1 \times c_{1}}$ denotes the prediction of unlabeled target data on the UDA task, which is transformed from the distribution of the pre-trained head. In Eq. (13), the discrepancy between the distribution $p_{g}^{t}$ and the prototype $M$ is described by the KL divergence and converted into a probability distribution using softmax. 

When the target data output of the task head $p_{h}^{t}$ is similar to the transformed probability distribution $\tilde{p}_{h}^{t}$, the generated pseudo-label is more reliable, allowing for more intense training. Leveraging this characteristic, we propose the calibration factor ${{\beta }^{t}}$ for Gini impurity, which is defined as follows:
\begin{eqnarray}
	\begin{aligned}
		\beta^{t}=\exp \left(-\mathrm{KL}\left(\tilde{p}_{h}^{t} \| p_{h}^{t}\right)\right).
	\end{aligned}
\end{eqnarray}
where ${{\beta }^{t}}\in \left( 0,1 \right]$. Hence, the calibrated Gini impurity is formulated as:
\begin{eqnarray}
	\begin{aligned}
		\mathcal{L}_{\text {cgi }}=\beta^{t} \mathrm{GI}\left(p_{h}^{t}\right).
	\end{aligned}
\end{eqnarray}

However, when the value of ${{\beta }^{t}}$ is tiny, the pseudo-label learning for the sample will stagnate. To address this issue, we modify  ${{\mathcal{L}}_{\text{cgi}}}$ as follows:
\begin{eqnarray}
	\begin{aligned}
		\mathcal{L}_{\mathrm{cgi}}=\beta^{t} \mathrm{G} \mathrm{I}\left(p_{h}^{t}\right)+\left(1-\beta^{t}\right) \mathrm{G} \mathrm{I}\left(p_{m}^{t}\right),
	\end{aligned}
\end{eqnarray}
\begin{eqnarray}
	\begin{aligned}
		p_{m}^{t}=0.5\left(\tilde{p}_{h}^{t}+p_{h}^{t}\right).
	\end{aligned}
\end{eqnarray}

From Eq. (16), it is shown that the probability distribution of the task head will be guided by the mixed probabilities $p_{m}^{t}$ for pseudo-label learning when the prediction $p_{h}^{t}$ is differs transformed probability $\tilde{p}_{h}^{t}$.

Given that the pseudo-label learning process introduces noise, the CGI only updates the task head ${{\theta }_{h}}$ without updating the feature extractor $\theta $. The following provides a brief analysis of how CGI calibrates Gini impurity. During training, ${{\beta }^{t}}$ and $\tilde{p}_{h}^{t}$ are treated as constants and do not contribute to the gradient computation. Consequently, the gradient of the CGI can be easily derived as:
\begin{eqnarray}
	\begin{aligned}
		\frac{\partial \mathcal{L}_{\mathrm{cgi}}}{\partial \theta_{h}}=-\mathbb{E}_{p_{t}}\left[\left(\left(1+\beta^{t}\right) p_{h}^{t}+\left(1-\beta^{t}\right) \tilde{p}_{h}^{t}\right) \frac{\partial p_{h}^{t}}{\partial \theta_{h}}\right].
	\end{aligned}
\end{eqnarray}

For convenience, the gradient is expressed in the expected form. From Eq. (18), it can be observed that the transformed probability will serve as a scaling factor to calibrate the gradient, along with the calibration factor. Additionally, it is worth noting that CGI will degrade to GI when the calibration factor equals 1.

\begin{algorithm}[t]
	\caption{Pseudo-code of BiPC.}
	\label{alg:1}
	\hspace*{0.02in} {\bf Input:}
	feature extractor ${F}_{\theta }$, pre-trained head ${G}_{{\theta }_{g}}$, task head $ {H}_{{{\theta }_{h}}}$, source dataset $D_s$, unlabeled target dataset $D_t$, learning rate $\eta$.
	\begin{algorithmic}[1]
		\State  Divide $D_s$ in 1:1 and learn the prototype $M$ with category relationship learning algorithm (\cite{ref18}).
		\For{$t=1...MaxIter$}
		\State \textbf{Sample} a batch data from $D_s$ and $D_t$.
		\State \textbf{Obtain} probability distribution from $ {H}_{{{\theta }_{h}}}$:
		\State $p_{h}^{s}={{H}_{{{\theta }_{h}}}}\left( {{F}_{\theta }}\left( {{x}^{s}} \right) \right), \textbf{ } p_{h}^{t}={{H}_{{{\theta }_{h}}}}\left( {{F}_{\theta }}\left( {{x}^{t}} \right) \right).$
		\State \textbf{Obtain} probability distribution from ${G}_{{\theta }_{g}}$:
		\State $p_{g}^{s}={{G}_{{{\theta }_{g}}}}\left( {{F}_{\theta }}\left( {{x}^{s}} \right) \right), \text{ } p_{g}^{t}={{G}_{{{\theta }_{g}}}}\left( {{F}_{\theta }}\left( {{x}^{t}} \right) \right).$
		\State \textbf{Calculate} classification loss ${{\mathcal{L}}_{\text{cls}}}$ with $p_{h}^{s}$ and ${y}^{s}$.
		\State \textbf{Obtain} ${{\mathcal{L}}_{cpa}}$ with $\left( p_{g}^{s},p_{g}^{t},{{y}^{s}},p_{h}^{t},M \right)$ by Eq. (11).
		\State \textbf{Obtain} ${{\mathcal{L}}_{cgi}}$ with $\left( p_{h}^{t},p_{g}^{t},M \right)$ by Eq. (16).
		\State \textbf{Update} ${F}_{\theta }$ with minimizing $\mathcal{L}_{\text{cls}}$ and $\mathcal{L}_{\text{cpa}}$:
		\State \hspace{1.0cm} $\theta \leftarrow \theta -\eta {{\nabla }_{\theta }}\left[ {{\mathcal{L}}_{cls}}+{{\mathcal{L}}_{cpa}} \right].$
		\State \textbf{Minimize} $\mathcal{L}_{\text{cls}}$ and $\mathcal{L}_{\text{cgi}}$ to update task head ${{H}_{{{\theta }_{h}}}}$:
		\State \hspace{1.0cm} ${{\theta }_{h}}\leftarrow {{\theta }_{h}}-\eta {{\nabla }_{{{\theta }_{h}}}}\left[ {{\mathcal{L}}_{cls}}+{{\mathcal{L}}_{cgi}} \right].$
		\State \textbf{Update} pre-trained head ${{G}_{{{\theta }_{g}}}}$ optimizing $\mathcal{L}_{\text{cpa}}$:
		\State \hspace{1.0cm} ${{\theta }_{g}}\leftarrow {{\theta }_{g}}-\eta {{\nabla }_{{{\theta }_{g}}}}\left[ {{\mathcal{L}}_{cpa}} \right].$
		\EndFor
	\end{algorithmic}
\end{algorithm}

\subsection{Bidirectional Probability Calibration}
Based on the calibrated probability alignment (Section 3.1) and calibrated Gini impurity (Section 3.2), we propose the BiPC framework for the UDA task. In the BiPC framework, the two tasks aim to learn calibration factors that enhance performance by leveraging information from their counterparts. Moreover, this method is easy to implement and can be integrated into various network architectures. The pseudo-code for the BiPC framework is provided in Algorithm 1.

\section{Experiments}
We evaluate BiPC against competitive UDA baseline on popular benchmarks. These datasets include \textbf{Office-Home}, \textbf{Office-31}, \textbf{Visda-2017} and \textbf{ImageCLEF-DA}. Apart from the datasets, digits classification constructed from \textbf{MNIST}, \textbf{USPS} and \textbf{SVHN} is taken into account. Besides, in ablation studies, BiPC is shown to improve adaption for different network architectures and the effect of each component is investigated.

\subsection{Setup}
\textbf{Office-Home} (\cite{ref21}) contains 15,588 images, which consists of images from 4 different domains: Artistic images (Ar), Clip Art (Cl), Product images (Pr) and Real-World images (Re). Collected in office and home settings, the dataset contains images of 65 object categories for each domain. We use all domain combinations and construct 12 transfer tasks.

\textbf{Office-31} (\cite{ref42}) is a benchmark dataset for domain adaptation which collected from three distinct domains: Amazon (A), Webcam (W) and DSLR (D). The dataset comprises 4,110 images in 31 classes, taken by web camera and digital SLR camera with different photographical settings, respectively. 6 transfer tasks are performed to enable unbiased evaluation.

\textbf{VisDA-2017} (\cite{ref43}) is a difficult simulation-to-real dataset with two very separate domains: Synthetic, renderings of 3D models from various perspectives and under various lighting conditions; Real, natural images. Over 280K photos from 12 different classes make up its training, validation and test domains.

\textbf{Digits} is a common UDA benchmark for digit recognition, utilizing three subsets: SVHN (S) (\cite{ref44}), MNIST (M) (\cite{ref45}) and USPS (U) (\cite{ref46}). We use the training sets to train our model and publish the recognition results using the standard test set from the target domain.

\textbf{ImageCLEF-DA} is a benchmark dataset for the ImageCLEF 2014 domain adaptation challenge, constructed by picking 12 common categories shared by the three public datasets, each of which is designated a domain: Caltech-256 (C), ImageNet ILSVRC 2012 (I) and Pascal VOC 2012. (P). Each category has 50 photographs and each domain has 600 images. We develop six transfer tasks using all domain combinations.

\textbf{Baseline Methods}. Our method will be compared against the state-of-the-art on different datasets, as outlined below:

\textit{Office-Home.} DSAN (\cite{ref9}), SHOT (\cite{ref58}), TOCL (\cite{ref78}), HOMDA (\cite{ref79}), SPL (\cite{ref80}), DTR (\cite{ref81}), CGDM (\cite{ref60}), SHOT (\cite{ref58}), CDTrans-B (\cite{ref14}), TVT (\cite{ref15}) and SSRT (\cite{ref16}).

\textit{Office-31.} DANN (\cite{ref5}), DSAN (\cite{ref9}), SHOT (\cite{ref58}), CAGAN (\cite{ref77}), TOCL (\cite{ref78}), CAGAN (\cite{ref77}), HOMDA (\cite{ref79}), SPL (\cite{ref80}), DTR (\cite{ref81}), CGDM (\cite{ref60}), CDTrans-B (\cite{ref14}), TVT (\cite{ref15}) and SSRT (\cite{ref16}).

\textit{VisDA-2017.} SHOT (\cite{ref58}), CDAN+MCC (\cite{ref67}), DSAN (\cite{ref9}), DSAN (\cite{ref9}), CGDM (\cite{ref60}), CDAN+E (\cite{ref48}), DTR (\cite{ref81}), CDTrans (\cite{ref14}), TVT (\cite{ref15}) and SSRT (\cite{ref16}).

\textit{Digits.} ADDA (\cite{ref66}), ADR (\cite{ref61}), CDAN+E (\cite{ref48}), CyCADA (\cite{ref62}), SWD \cite{ref53}, rRGrad+CAT (\cite{ref51}), DSAN (\cite{ref9}), SHOT (\cite{ref58}) and HOMDA (\cite{ref79}).

\textit{ImageCLEF-DA.} DAN (\cite{ref4}), DANN (\cite{ref5}), D-CORAL (\cite{ref6}), MADA (\cite{ref32}), CDAN (\cite{ref48}), CDAN+E (\cite{ref48}) and DSAN (\cite{ref9}).

\textbf{Notation $\&$ Arrangement}. For a fair comparison, the following architectures will be used in the experiments: ResNet-50 (\cite{ref19}), ResNet-101 (\cite{ref19}), AlexNet (\cite{ref63}), SwinT (\cite{ref13}) and DeiT (\cite{ref64}). These architectures have been widely employed in related methods. In the ablation study, additional network architectures will be investigated, including ConvNeXt (\cite{ref68}), CrossViT (\cite{ref71}) and gMLP (\cite{ref20}), to assess the flexibility and effectiveness of our approach. For convenience, the mentioned network is abbreviated to the following form: ``ResNet-50/ResNet-101'' to ``R50/R101'', ``DeiT-Base'' to ``DB'', ``gMLP-S'' to ``gMS'', ``ConvNeXt-Base'' to ``CB'', ``CrossViT-Base'' to ``CVB'', ``SwinT-Base'' to ``SB'' and others as usual. An ``o'' indicates a model pre-trained on ImageNet-21k (\cite{ref65}) while no indication means pre-training was done on ImageNet-1k (\cite{ref17}). An ``*'' denotes the results from the literature. The term ``Baseline'' refers to training a backbone directly on the source domain and testing it on the target domain. Each cell in the experiment table represents the accuracy of domain migration from the source domain to the target domain, indicated as ``Source→Target'' or ``Synthesis→Real.'' The last column of the table represents the average accuracy across all subtasks, denoted as ``Avg.'' The network structure used for the comparison method is specified in the upper left corner of the table.

\textbf{Implementation Details}. For all tasks, mini-batch SGD with momentum of 0.9 and the weight decay ratio 5e-4 are adopted to optimize the training process. The annealing strategy is employed, formulated as $\eta =\frac{{{\eta }_{\text{0}}}}{{{(1+\tau \rho )}^{\upsilon }}}$, where $\rho$ is the training progress, $\tau=\text{3e-4}$ and $\upsilon=0.75$. ${{\eta }_{0}}$ is the initial learning rate, for the pre-trained part ${{F}_{\theta }}(\cdot)$ and ${{G}_{{{\theta }_{g}}}}(\cdot)$, set to 3e-4 for the Office-Home, Office-31 and Digits, 3e-5 for VisDA-2017. The learning rate of the task head ${{H}_{{{\theta }_{h}}}}(\cdot)$ is set to 10 times the pre-training part. The classification loss is performed with standard cross-entropy loss in addition to VisDA-2017 with focal loss (\cite{ref69}) since it can easily converge on the source domain. The tradeoff ${{\lambda }_{1}}$ is set to 1, while ${{\lambda }_{2}}$ and ${{\lambda }_{3}}$ adopt dynamic mechanism like DSAN (\cite{ref9}) to suppress noisy activations at the early stages of training: $\lambda =\frac{2 a}{\exp \left( -\delta \rho \right)}-1$ and $\delta =10$. $a$ of ${{\lambda }_{2}}$ is set to 1 and of ${{\lambda }_{3}}$ to 0.25. The batch size is set to 32, except for the Digits task, where it is set to 256. The total number of epochs is set to 20, and the label smoothing factor is set to 0.1. All experiments were conducted using an NVIDIA GTX 3090 with 24G RAM, and the PyTorch framework was used to implement our experiments.

% Table1
\begin{table*}[t]
	\scriptsize
	\setlength{\abovecaptionskip}{0cm}  %段前
	\setlength{\belowcaptionskip}{-0.0cm} %段后
	\centering
	\caption{Comparison with SoTA methods on Office-Home for vanilla closed-set UDA. ``*'' indicate the results from the literature that follows. ``o'' implies its pre-trained from on ImageNet-21K instead of ImageNet-1K. The best performance is marked as bold.}
	\renewcommand{\arraystretch}{1.1}
	\setlength{\tabcolsep}{1.5mm}{
		\begin{tabular}{c|ccccccccccccc}
			\toprule
			Method (Source→Target)& {Ar→Cl} & {Ar→Pr} & {Ar→Re} & {Cl→Ar} & {Cl→Pr} & {Cl→Re} & {Pr→Ar} & {Pr→Cl} & {Pr→Re} & {Re→Ar} & {Re→Cl} & {Re→Pr} & \cellcolor{gray!25}{Avg.} \\
			\midrule
			\hspace{-0.7cm}\makecell[l]{\textit{ResNet50:}} &       &       &       &       &       &       &       &       &       &       &       &       &\cellcolor{gray!25}  \\
			DSAN (\cite{ref9})	&54.4	&70.8	&75.4	&60.4	&67.8	&68.0	&62.6	&55.9	&78.5	&73.8	&60.6	&83.1	&\cellcolor{gray!25}67.6\\
			SHOT (\cite{ref58})	&57.1	&78.1	&81.5	&68.0	&78.2	&78.1	&67.4	&54.9	&\textbf{82.2}	&73.3	&58.8	&84.3	&\cellcolor{gray!25}71.8\\
			TOCL (\cite{ref78})	&54.8	&73.9	&79.5	&63.3	&73.2	&75.8	&63.6	&55.0	&80.2	&73.8	&58.4	&\textbf{85.3}	& \cellcolor{gray!25}69.2\\
			{HOMDA (\cite{ref79})}	&57.3	&73.7	&\textbf{84.2}	&64.1	&73.8	&76.4	&64.8	&55.5	&82.2	&71.3	&58.6	&79.9	& \cellcolor{gray!25}70.2\\
			{SPL (\cite{ref80})} 	&51.6	&76.0	&80.6	&60.3	&77.0	&78.4	&62.9	&50.7	&81.2	&66.3	&52.8	&82.9	& \cellcolor{gray!25}68.6\\
			{DTR (\cite{ref81})}	&56.4	&76.5	&79.1	&63.7	&75.1	&74.9	&65.5	&56.2	&80.7	&75.5	&\textbf{61.0}	&\textbf{85.3}	&\cellcolor{gray!25}70.9\\
			Baseline-R50	&51.0	&68.2	&74.8	&54.2	&63.6	&66.8	&53.6	&45.4	&74.5	&65.6	&53.5	&79.3	& \cellcolor{gray!25}62.5\\
			Ours-R50	&\textbf{59.1}	&\textbf{79.3}	&82.2	&\textbf{68.3}	&\textbf{78.4}	&\textbf{79.0}	&\textbf{67.3}	&\textbf{57.1}	&81.7	&\textbf{74.0}	&60.3	&84.8	&\cellcolor{gray!25}\textbf{72.6}\\
			\midrule
			\hspace{-1.0cm}\makecell[l]{\textit{ViT:}} &       &       &       &       &       &       &       &       &       &       &       &       & \cellcolor{gray!25} \\
			CGDM-B$^*$ (\cite{ref14})	&67.1	&83.9	&85.4	&77.2	&83.3	&83.7	&74.6	&64.7	&85.6	&79.3	&69.5	&87.7	&\cellcolor{gray!25}78.5\\
			SHOT-B$^*$ (\cite{ref14})	&67.1	&83.5	&85.5	&76.6	&83.4	&83.7	&76.3	&65.3	&85.3	&80.4	&66.7	&83.4	&\cellcolor{gray!25}78.1\\
			CDTrans-B (\cite{ref14})	&68.8	&85.0	&86.9	&81.5	&87.1	&87.3	&\textbf{79.6}	&63.3	&88.2	&\textbf{82.0}	&66.0	&90.6	&\cellcolor{gray!25}80.5\\
			Baseline-DB	&61.8	&79.5	&84.3	&75.4	&78.8	&81.2	&72.8	&55.7	&84.4	&78.3	&59.3	&86.0	&\cellcolor{gray!25}74.8\\
			Ours-DB	&\textbf{71.0}	&\textbf{86.5}	&\textbf{89.3}	&\textbf{81.7}	&\textbf{87.3}	&\textbf{87.5}	&79.3	&\textbf{66.7}	&\textbf{88.6}	&81.3	&\textbf{69.1}	&\textbf{91.0}	&\cellcolor{gray!25}\textbf{81.6}\\
			\midrule
			TVT$^o$  (\cite{ref15})	&74.9	&86.8	&89.4	&82.8	&88.0	&88.3	&79.8	&71.9	&90.1	&85.5	&74.6	&90.6	&\cellcolor{gray!25}83.6\\
			SSRT-B$^o$ (\cite{ref16})	&\textbf{75.2}	&89.0	&91.1	&85.1	&88.3	&90.0	&85.0	&\textbf{74.2}	&91.3	&85.7	&\textbf{78.6}	&91.8	&\cellcolor{gray!25}85.4\\
			Baseline-DB$^o$ 	&70.0	&86.2	&89.3	&81.6	&85.6	&87.8	&79.9	&68.2	&89.3	&82.2	&69.5	&89.2	&\cellcolor{gray!25}81.6\\
			Ours-DB$^o$	&74.9	&\textbf{90.1}	&\textbf{91.3}	&\textbf{86.3}	&\textbf{91.0}	&\textbf{91.1}	&\textbf{85.5}	&71.3	&\textbf{91.7}	&\textbf{86.7}	&73.6	&\textbf{92.6}	&\cellcolor{gray!25}\textbf{85.5}\\
			\midrule
			\hspace{-0.7cm}\makecell[l]{\textit{SwinT:}} &       &       &       &       &       &       &       &       &       &       &       &       & \cellcolor{gray!25} \\
			Baseline-SB$^o$	&72.1&	85.6&	89.3&	84.4&	87.6&	89.2&	82.9&	72.0&	90.0&	84.0&	73.3&	91.0   &\cellcolor{gray!25}83.4\\
			Ours-SB$^o$	&\textbf{79.1} &	\textbf{91.0}&	\textbf{92.5}&	\textbf{88.1}&	\textbf{91.5}&	\textbf{91.5}&	\textbf{86.4}&	\textbf{77.8} &	\textbf{92.3} &	\textbf{87.1} & \textbf{77.5} &	\textbf{92.3}&	\cellcolor{gray!25}\textbf{86.7}\\
			\bottomrule
		\end{tabular}%
	}
	\label{table2}%
\end{table*}%
% Table generated by Excel2LaTeX from sheet 'Table2'
\begin{table}[htbp]
	\scriptsize
	\setlength{\abovecaptionskip}{0cm}  %段前
	\setlength{\belowcaptionskip}{-0.0cm} %段后
	\centering
	\caption{Comparison with SoTA methods on Office-31 for vanilla closed-set UDA.}
	\renewcommand{\arraystretch}{1.1}
	\setlength{\tabcolsep}{0.3mm}{
		\begin{tabular}{c|ccccccc}
			\toprule
			Method (Source→Target) & A→D & A→W & D→A & D→W & W→A & W→D & \cellcolor{gray!25}Avg. \\
			\midrule
			\hspace{-0.7cm}\makecell[l]{\textit{ResNet50:}} &       &       &       &       &       &       & \cellcolor{gray!25} \\
			DANN$^*$ (\cite{ref9}) & 79.7  & 82.0  & 68.2  & 96.9  & 67.4  & 99.1  & \cellcolor{gray!25}82.2  \\
			{DSAN (\cite{ref9})} & 90.2  & 93.6  & 73.5  & 98.3  & 74.8  & \textbf{100.0} & \cellcolor{gray!25}88.4  \\
			{SHOT (\cite{ref58})} & 94.0  & 90.1  & 74.7  & 98.7 & 74.3  & 99.8  & \cellcolor{gray!25}88.6  \\
			CAGAN (\cite{ref77}) & 93.2  & 93.8  & 74.5  & 98.1  & 75.2  & \textbf{100.0} & \cellcolor{gray!25}89.1  \\
			TOCL (\cite{ref78}) & 92.6  & 92.7  & 73.5  & 99.2  & 72.2  & \textbf{100.0} & \cellcolor{gray!25}88.4  \\
			{HOMDA (\cite{ref79})} & 93.9  & 93.6  & 75.8  & 98.7  & 75.4  & 99.8  & \cellcolor{gray!25}89.5  \\
			{SPL (\cite{ref80})} & 93.0  & 88.6  & 73.8  & 98.7  & 74.2  & 97.1 & \cellcolor{gray!25}87.6  \\
			{DTR (\cite{ref81})} & 94.0  & \textbf{95.6} & 76.4  & \textbf{99.0} & 75.6  & \textbf{100.0} & \cellcolor{gray!25}\textbf{90.2}  \\
			Baseline-R50 & 80.6  & 97.6  & 66.1  & 82.7  & 65.0  & \textbf{100.0} & \cellcolor{gray!25}82.0  \\
			{Ours-R50} & \textbf{94.6} & 95.4  & \textbf{76.8} & 97.7  & \textbf{76.9} & \textbf{100.0} & \cellcolor{gray!25}\textbf{90.2} \\
			\midrule
			\hspace{-1.0cm}\makecell[l]{\textit{ViT:}} &       &       &       &       &       &       & \cellcolor{gray!25} \\
			{CGDM-B$^*$ (\cite{ref14})} & 94.6  & 95.3  & 78.8  & 97.6  & 81.2  & 99.8  & \cellcolor{gray!25}91.2  \\
			{SHOT-DB$^*$ (\cite{ref14})} & 95.3  & 94.3  & 79.4  & 99.0  & 80.2  & \textbf{100.0} & \cellcolor{gray!25}91.4  \\
			{CDTrans-DB (\cite{ref14})} & 97.0  & 96.7  & 81.1  & 99.0  & 81.9  & \textbf{100.0} & \cellcolor{gray!25}92.6  \\
			{Baseline-DB} & 90.8  & 90.4  & 76.8  & 98.2  & 76.4  & \textbf{100.0} & \cellcolor{gray!25}88.8  \\
			{Ours-DB} & \textbf{97.3} & \textbf{97.3} & \textbf{81.7} & \textbf{99.1} & \textbf{82.3} & \textbf{100.0} & \cellcolor{gray!25}\textbf{92.9} \\
			\midrule
			{TVT$^o$ (\cite{ref15})} & 96.4  & 96.4  & \textbf{84.9} & \textbf{99.4} & \textbf{86.1} & \textbf{100.0} & \cellcolor{gray!25}93.8  \\
			{SSRT-B$^o$ (\cite{ref16})} & \textbf{98.6} & 97.7  & 83.5  & 99.2  & 82.2  & \textbf{100.0} & \cellcolor{gray!25}93.5  \\
			{Baseline-DB$^o$} & 95.4  & 95.3  & 83.5  & 99.1  & 84.1  & \textbf{100.0} & \cellcolor{gray!25}92.9  \\
			{Ours-DB$^o$} & 98.2  & \textbf{98.7} & 84.2  & 99.2  & 84.8  & \textbf{100.0} & \cellcolor{gray!25}\textbf{94.2} \\
			\midrule
			\hspace{-0.7cm}\makecell[l]{\textit{{SwinT:}}} &       &       &       &       &       &       &  \cellcolor{gray!25}\\
			{Baseline-SB$^o$} & 94.1  & 94.9  & 82.6  & 99.0  & 83.1  & 100.0  & \cellcolor{gray!25}92.3  \\
			{Ours-SB$^o$} & \textbf{98.6} & \textbf{99.1} & \textbf{84.7} & \textbf{99.2} & \textbf{86.1} & \textbf{100.0} & \cellcolor{gray!25}\textbf{94.0} \\
			\bottomrule
		\end{tabular}%
	}
	\label{table3}%
\end{table}%

\begin{table*}[htbp]
	\scriptsize
	\centering
	\caption{Comparison with SoTA methods on VisDA-2017 for vanilla closed-set UDA.}
	\renewcommand{\arraystretch}{1.1}
	\begin{tabular}{c|ccccccccccccc}
		\toprule
		Method (Synthesis→Real)& plane & {bcycl} & {bus} & {car} & {horse} & {knife} & {mcycl} & {person} & {plant} & {sktbrd} & {train} & {truck} & \cellcolor{gray!25}{Avg.} \\
		\midrule
		\hspace{-0.7cm}\makecell[l]{\textit{ResNet101:}} &       &       &       &       &       &       &       &       &       &       &       &       & \cellcolor{gray!25} \\
		SHOT (\cite{ref58}) & \textbf{95.5}  & \textbf{87.5} & 80.1  & 54.5  & 93.6  & 94.2  & 80.2  & 80.9  & 90.0  & 89.9 & 87.1  & \textbf{58.4} & \cellcolor{gray!25}82.7  \\
		CDAN+MCC (\cite{ref67}) & 94.5  & 80.8  & 78.4  & 65.3  & 90.6  & 79.4  & 87.5  & 82.2  & 94.7  & 81.0  & 86.0  & 44.6  & \cellcolor{gray!25}80.4  \\
		DSAN (\cite{ref9}) & 90.9  & 66.9  & 75.7  & 62.4  & 88.9  & 77.0  & 93.7 & 75.1  & 92.8  & 67.6  & 89.1 & 39.4  & \cellcolor{gray!25}75.1  \\
		CGDM$^*$ (\cite{ref14}) & 92.8  & 85.1  & 76.3  & 64.5  & 91.0  & 93.2  & 81.3  & 79.3  & 92.4  & 83.0  & 85.6  & 44.8  & \cellcolor{gray!25}80.8  \\
		CDAN+E$^*$ (\cite{ref14}) & 85.2  & 66.9  & 83.0  & 50.8  & 84.2  & 74.9  & 88.1  & 74.5  & 83.4  & 76.0  & 81.9  & 38.0  & \cellcolor{gray!25}73.9  \\
		DTR$^*$ (\cite{ref81}) & 94.2  & 75.1  & 80.2  & 66.1  & 92.1  & 89.6  & \textbf{94.9}  & 80.8  & 90.9  & 76.4  & \textbf{89.2}  & 50.3  & \cellcolor{gray!25}81.7  \\
		Baseline-R101 & 63.0  & 34.6  & 30.3  & 48.1  & 69.5  & 13.5  & 91.7  & 23.2  & 66.1  & 27.8  & 77.2  & 6.4   & \cellcolor{gray!25}46.0  \\
		Ours-R101 & 91.1  & 81.4  & 79.7  & 66.5  & 92.5  & 94.6  & 91.4  & 77.3  & 93.2  & 71.6  & 88.3  & 50.5  & \cellcolor{gray!25}81.7  \\
		Ours-R101+FixMatch & 94.1  & 83.3  & \textbf{83.3}  & \textbf{74.6} & \textbf{95.9} & \textbf{95.7} & 91.5  & \textbf{82.3} & \textbf{95.0} & \textbf{77.6}  & 87.7  & 48.7  & \cellcolor{gray!25}\textbf{84.1} \\
		\midrule
		\hspace{-1.2cm}\makecell[l]{\textit{ViT:}} &       &       &       &       &       &       &       &       &       &       &       &       &  \cellcolor{gray!25}\\
		CGDM-B$^*$ (\cite{ref14}) & 96.3  & 87.1  & 86.8  & \textbf{83.5} & 92.2  & 98.3  & 91.6  & 78.5  & 96.3  & 48.4  & 89.4  & 39.0  & \cellcolor{gray!25}82.3  \\
		SHOT-B$^*$ (\cite{ref14}) & 97.9  & 90.3  & 86.0  & 73.4  & 96.9  & \textbf{98.8} & 94.3  & 54.8  & 95.4  & 87.1  & 93.4  & 62.7  & \cellcolor{gray!25}85.9  \\
		CDTrans-B (\cite{ref14}) & 97.1  & \textbf{90.5} & 82.4  & 77.5  & 96.6  & 96.1  & 93.6  & \textbf{88.6} & 97.9  & 86.9  & 90.3  & \textbf{62.8} & \cellcolor{gray!25}88.4  \\
		Baseline-DB & 97.9  & 80.8  & 76.8  & 66.7  & 77.6  & 63.3  & 95.8  & 9.3   & 96.0  & 41.6  & 90.4  & 7.7   & \cellcolor{gray!25}67.0  \\
		Ours-DB & 98.4  & 88.9  & 86.5  & 74.3  & 97.0  & 96.6  & 96.2  & 70.3  & 97.2  & 88.2  & 92.3  & 58.7  & \cellcolor{gray!25}87.1  \\
		Ours-DB+FixMatch & \textbf{98.6} & 90.1  & \textbf{91.2} & 82.0  & \textbf{98.1} & 97.3  & \textbf{97.3} & 76.0  & \textbf{98.1} & \textbf{90.7} & \textbf{94.0} & 50.0  & \cellcolor{gray!25}\textbf{88.6} \\
		\midrule
		TVT$^o$ (\cite{ref15}) & 92.9  & 85.6  & 77.5  & 60.5  & 93.6  & 98.2  & 89.4  & 76.4  & 93.6  & 92.0  & 91.7  & 55.7  & \cellcolor{gray!25}83.9  \\
		SSRT-B$^o$ (\cite{ref16}) & 98.9  & 87.6  & 89.1  & \textbf{84.7} & 98.3  & \textbf{98.7} & \textbf{96.2} & \textbf{81.0} & 94.8  & \textbf{97.9} & 94.5  & 43.1  & \cellcolor{gray!25}\textbf{88.7} \\
		Baseline-DB$^o$ & \textbf{99.2} & 62.2  & 91.6  & 72.7  & 95.7  & 87.4  & 93.3  & 10.9  & 91.2  & 80.9  & 94.3  & 29.0  & \cellcolor{gray!25}75.7  \\
		Ours-DB$^o$ & 98.3  & 86.1  & 87.3  & 67.3  & 96.4  & 92.8  & 93.5  & 65.5  & 95.4  & 89.6  & 93.5  & \textbf{66.8} & \cellcolor{gray!25}86.0  \\
		Ours-DB$^o$+FixMatch & 98.7  & \textbf{88.1} & 90.8  & 76.4  & \textbf{98.5} & 96.0  & 95.0  & 66.1  & \textbf{96.2} & 91.6  & \textbf{95.1} & 58.9  & \cellcolor{gray!25}87.6  \\
		\midrule
		\hspace{-1.0cm}\makecell[l]{\textit{SwinT:}} &       &       &       &       &       &       &       &       &       &       &       &       &  \cellcolor{gray!25}\\
		Baseline-SB$^o$ & \textbf{99.6} & 67.7  & 85.7  & 62.7  & \textbf{98.7} & 82.1  & \textbf{98.1} & 28.0  & 88.4  & 90.7  & \textbf{97.1} & 30.5  & \cellcolor{gray!25}77.4  \\
		Ours-SB$^o$ & 99.5  & 90.9  & 87.2  & 73.3  & 98.6  & 97.5  & 97.2  & \textbf{73.6} & 96.9  & 95.7  & 94.1  & \textbf{55.5} & \cellcolor{gray!25}88.3  \\
		Ours-SB$^o$+FixMatch & 99.4  & \textbf{91.5} & \textbf{90.7} & \textbf{81.1} & \textbf{98.7} & \textbf{98.1} & 97.2  & 71.6  & \textbf{97.7} & \textbf{98.4} & 95.5  & 51.2  & \cellcolor{gray!25}\textbf{89.3} \\
		\bottomrule
	\end{tabular}%
	\label{table4}%
\end{table*}%

% Table1
% Table generated by Excel2LaTeX from sheet 'Table5'
\begin{table}[h]
	\scriptsize
	\setlength{\abovecaptionskip}{0cm}  %段前
	\setlength{\belowcaptionskip}{-0.0cm} %段后
	\centering
	\caption{Comparison with SoTA methods on Digits for vanilla closed-set UDA.}
	\renewcommand{\arraystretch}{1.1}
	\setlength{\tabcolsep}{1.5mm}{
		\begin{tabular}{c|cccc}
			\toprule
			Method (Source→Target) & {S→M} & {U→M} & {M→U} & \cellcolor{gray!25}{Avg.} \\
			\midrule
			\hspace{-1.0cm}\makecell[l]{\textit{LetNet5:}} &       &       &       & \cellcolor{gray!25} \\
			ADDA (\cite{ref66}) & 76.0  & 90.1  & 89.4  & \cellcolor{gray!25}85.2  \\
			ADR (\cite{ref61}) & 95.0  & 93.1  & 93.2  & \cellcolor{gray!25}93.8  \\
			CDAN+E (\cite{ref48}) & 89.2  & 98.0  & 95.6  & \cellcolor{gray!25}94.3  \\
			CyCADA (\cite{ref62}) & 90.4  & 96.5  & 95.6  & \cellcolor{gray!25}94.2  \\
			SWD (\cite{ref53}) & 98.9  & 97.1  & 98.1 & \cellcolor{gray!25}98.0  \\
			rRGrad+CAT (\cite{ref51}) & 98.8  & 96.0  & 94.0  & \cellcolor{gray!25}96.3  \\
			DSAN (\cite{ref9}) & 90.1  & 95.3  & 96.9  & \cellcolor{gray!25}94.1  \\
			SHOT (\cite{ref58}) & \textbf{99.0} &97.6 &97.8 &\cellcolor{gray!25}98.1 \\
			{HOMDA (\cite{ref79})} & \textbf{98.6} &97.5 &98.6 &\cellcolor{gray!25}98.1 \\
			\midrule
			\hspace{-1.0cm}\makecell[l]{\textit{AlexNet:}} &       &       &       & \cellcolor{gray!25} \\
			Baseline & 69.1  & 50.1  & 78.1  & \cellcolor{gray!25}65.7  \\
			Ours  & \textbf{98.9} & \textbf{98.2} & \textbf{97.5} & \cellcolor{gray!25}\textbf{98.2} \\
			\midrule
			\hspace{-1.0cm}\makecell[l]{\textit{ViT:}} &       &       &       & \cellcolor{gray!25} \\
			Baseline-DB & 85.5  & 89.4  & \textbf{97.1} & \cellcolor{gray!25}90.7  \\
			Ours-DB & \textbf{98.0} & \textbf{97.4} & \textbf{97.1} & \cellcolor{gray!25}\textbf{97.5} \\
			\midrule
			\hspace{-1.0cm}\makecell[l]{\textit{SwinT:}} &       &       &       &  \cellcolor{gray!25}\\
			Baseline-SB$^o$ & \textbf{86.5}  & \textbf{97.5}  & \textbf{96.5}  &  \cellcolor{gray!25}93.5  \\
			Ours-SB$^o$ & \textbf{96.1}  & \textbf{97.5}  & \textbf{96.5}   & \cellcolor{gray!25}\textbf{96.7}  \\
			\bottomrule
		\end{tabular}%
	}
	\label{table5}%
\end{table}%
% Table generated by Excel2LaTeX from sheet 'Table6'
\begin{table}[htp]
	\scriptsize
	\setlength{\abovecaptionskip}{0cm}  %段前
	\setlength{\belowcaptionskip}{-0.0cm} %段后
	\centering
	\caption{Comparison with SoTA methods on ImageCLEF-DA for vanilla closed-set UDA.}
	\renewcommand{\arraystretch}{1.1}
	\setlength{\tabcolsep}{0.5 mm}{
		\begin{tabular}{c|ccccccc}
			\toprule
			Method (Source→Target)& {I→P} & {I→C} & {P→I} & {P→C} & {C→I} &{C→P} & \cellcolor{gray!25}{Avg.} \\
			\midrule
			\hspace{-0.7cm}\makecell[l]{\textit{ResNet50:}} &       &       &       &       &       &       & \cellcolor{gray!25} \\
			DAN (\cite{ref4}) & 75.0  & 93.3  & 86.2  & 91.3  & 84.1  & 69.8  &\cellcolor{gray!25}83.3  \\
			DANN (\cite{ref5}) & 75.0  & 96.2  & 86.0  & 91.5  & 87.0  & 74.3  &\cellcolor{gray!25}85.0  \\
			D-CORAL (\cite{ref6}) & 76.9  & 93.6  & 88.5  & 91.6  & 86.8  & 74.0  &\cellcolor{gray!25}85.2  \\
			MADA (\cite{ref32}) & 75.0  & 96.0  & 87.9  & 92.2  & 88.8  & 75.2  &\cellcolor{gray!25}85.8  \\
			CDAN (\cite{ref48}) & 76.7  & 97.0  & 90.6  & 93.5  & 90.5  & 74.5  & \cellcolor{gray!25}87.1  \\
			CDAN+E (\cite{ref48}) & 77.7  & \textbf{97.7} & 90.7  & 94.3  & 91.3  & 74.2  & \cellcolor{gray!25}87.7  \\
			DSAN (\cite{ref9}) & \textbf{80.2} & 97.2  & 93.3  & 95.9  & 93.8  & \textbf{80.8} & \cellcolor{gray!25}\textbf{90.2} \\
			Baseline-R50 & 75.8  & 92.1  & 85.6  & 88.5  & 84.0  & 67.3  & \cellcolor{gray!25}82.2  \\
			Ours-R50 & 79.0  & 97.2  & \textbf{93.7} & \textbf{96.2} & \textbf{94.7} & 79.8  & \cellcolor{gray!25}90.1  \\
			\midrule
			\hspace{-0.7cm}\makecell[l]{\textit{ViT:}}  &       &       &       &       &       &       & \cellcolor{gray!25} \\
			Baseline-DB & 80.8  & 94.5  & 94.1  & 95.5  & 85.3  & 71.5  & \cellcolor{gray!25}86.9  \\
			Ours-DB & \textbf{86.3} & \textbf{99.5} & \textbf{96.3} & \textbf{98.0} & \textbf{96.7} & \textbf{81.8} & \cellcolor{gray!25}\textbf{93.1} \\
			\midrule
			\hspace{-0.7cm}\makecell[l]{\textit{SwinT:}}  &       &       &       &       &       &       & \cellcolor{gray!25} \\
			Baseline-SB$^o$ & 80.8   & 95.6   & 96.0   &94.7    &  92.5  & 74.7   & \cellcolor{gray!25}89.1   \\
			Ours-SB$^o$ & \textbf{84.2}   & \textbf{99.3}   & \textbf{98.3}   & \textbf{98.7}  & \textbf{98.5}  & \textbf{83.5}  &  \cellcolor{gray!25}\textbf{93.7} \\
			\bottomrule
		\end{tabular}%
	}
	\label{table6}%
\end{table}%

\subsection{Comparison to SoTA}
The results for Office-Home, Office-31, VisDA-2017, Digits, and ImageCLEF-DA can be found in Table 2, 3, 4, 5, and 6, respectively. We have conducted numerous experiments to ensure a fair comparison, evaluating our method against related approaches on different network structures, including CNNs and Transformers. Despite using similar hyperparameters across all network structures, our approach achieves comparable performance or even reaches the state-of-the-art on various backbones and datasets. These results demonstrate the efficiency and flexibility of the BiPC framework. Further results related to different network architectures are presented in the ``Further Network Architectures'' of Section 4.4

\textbf{Results on Office-Home}. Table 2 shows the quantitative results of methods. As expected, BiPC emerges as the SoTA in the ViT architecture, achieving an accuracy of 81.6\% and 85.5\% using the pre-trained model of ImageNet-1k and ImageNet-21k, respectively. Furthermore, BiPC achieves an impressive accuracy of 86.7\% under the SwinT-Base architecture. Under the ResNet-50 structure, BiPC achieves better performance (72.6\%) with the previous SoTA method (71.8\%).

\textbf{Results on Office-31}. Table 3 shows the quantitative results with the CNN-based and ViT-based methods. Overall, our method achieves the best performance on each task with 94.2\% average accuracy and outperforms the SoTA methods.

\textbf{Results on VisDA-2017}. As shown in Table 4, our BiPC achieves 89.3\% average accuracy and outperforms the baseline by +11.9\% with ``SwinT-Base+FixMatch''. Furthermore, the effectiveness of the method has been validated on different architectures, achieving accuracy of 81.5\% and 86.0\% on ResNet-50 and DeiT-Base. When combined with FixMatch, the performance can be further improved by +2.6\% and +1.6\%, respectively.

\textbf{Results on Digits}. As observed in Table 5, utilizing the AlexNet structure, our method achieves comparable performance (98.2\%) with the SoTA method (98.4\%), surpassing the Baseline by +32.5\%. 

\textbf{Results on ImageCLEF-DA}. Table 6 shows the quantitative results with the CNN-based and Transformer-based methods. Overall, BiPC achieves the best performance on each task with 93.7\% accuracy and outperforms the SoTA methods. Numerically, under the ResNet-50, Our method achieves a performance of 90.1\%, which is comparable to the state-of-the-art method that achieves 90.2\%.

The experimental results provide several insightful observations. Firstly, preliminary verification demonstrates that our method outperforms feature alignment. Comparing BiPC with category alignment methods, our method achieves superior performance. Secondly, for Transformer structures, feature alignment methods show limited improvements, while BiPC demonstrates better results. \cite{ref14} introduced various popular feature alignment methods for Transformers, but it is inferior to our approach in the same setting. Thirdly, in Transformer architectures, BiPC achieves comparable or even superior results compared to Transformer-based methods. Specifically, in the Office-Home and Office-31 datasets, BiPC outperforms Transformer-based approaches (\cite{ref14}), (\cite{ref15}), (\cite{ref16}) under the ViT structure, whether using ImageNet-1k or ImageNet-21k pre-trained models. Fourthly, the backbone architecture contributes more to performance improvement than the algorithms themselves. Our experiments show that Transformers alone can outperform CNN-based algorithms, and we further explore SwinT and find that it outperforms ViT-based UDA methods. This implies that instead of focusing solely on specific models for algorithm development, more attention should be given to developing general UDA algorithms. Fifthly, BiPC relies on pre-trained networks. Our method can be considered as an efficient way to leverage pre-trained models, utilizing the prior knowledge of the probability space to reduce the domain gap. However, BiPC is not applicable in the absence of pre-trained models. In the Digits dataset experiments, most relevant methods use LeNet-5 as a baseline, and we could not apply our method to the backbone due to the lack of a pre-trained model. Lastly, in the VisDA-2017 dataset, the advantage of BiPC is not evident due to the significant gap between the source and target domains. Considering that state-of-the-art methods implicitly incorporate data augmentation or regularization, we simply combine BiPC with FixMatch (\cite{ref70}), a classical semi-supervised algorithm. The results demonstrate that this simple combination yields effective improvements.

\begin{figure*}[htbp]
	\setlength{\abovecaptionskip}{0.cm}
	\setlength{\belowcaptionskip}{-0.cm}
	\centering
	\includegraphics[width=6in]{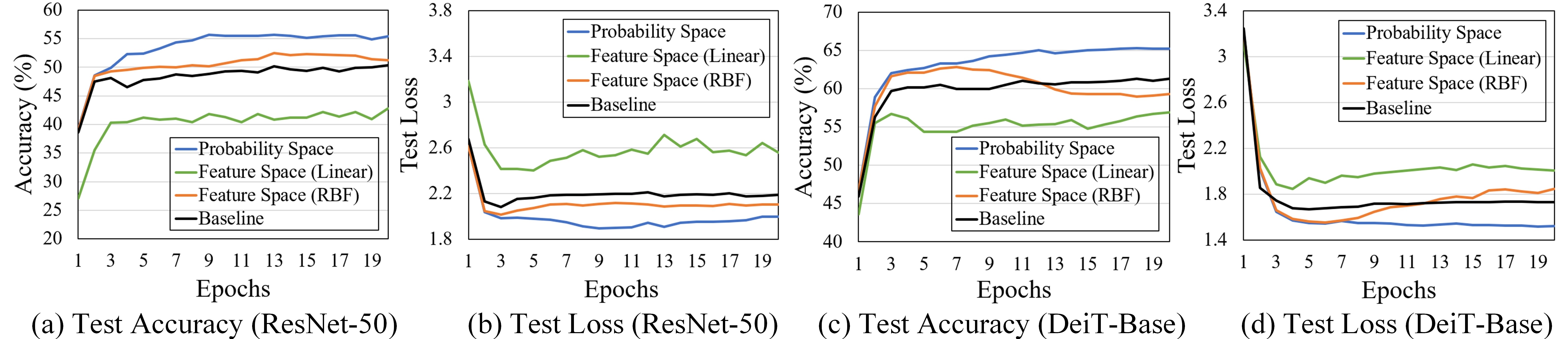}
	\caption{Convergence and performance of feature space vs probability space on task Art→Clipart (Office-Home). (a) Target domain accuracy with ResNet-50. (b) Target domain loss with ResNet-50. (c) Target domain accuracy with DeiT-Base. (d) Target domain loss with DeiT-Base.}
	\label{fig3}
\end{figure*}

% Table generated by Excel2LaTeX from sheet 'Table7'
\begin{table}[t]
	\scriptsize
	\setlength{\abovecaptionskip}{0cm}  %段前
	\setlength{\belowcaptionskip}{-0.0cm} %段后
	\renewcommand{\arraystretch}{1.1}
	\centering
	\caption{Contribution of each part on Office-Home with BiPC-R50.}
	\setlength{\tabcolsep}{4.0mm}{
		\begin{tabular}{cccc|cc}
			\toprule
			CLS & CPA & CGI ($\theta_{h}$) & CGI ($\theta_{h}$ $\theta$) & \cellcolor{gray!25}Avg. \\
			\midrule
			\ding{52}    &  &  &  &  \cellcolor{gray!25}62.5\\
			\ding{52}     & \ding{52}     &  &  &   \cellcolor{gray!25}68.4 \\
			\ding{52}     &  &  & \ding{52} &    \cellcolor{gray!25}70.6\\
			\ding{52}     & \ding{52}     &  & \ding{52} &  \cellcolor{gray!25}72.4  \\
			\ding{52}     & & \ding{52}    &  & \cellcolor{gray!25}69.6    \\
			\ding{52}     &\ding{52}     & \ding{52}     & & \cellcolor{gray!25}\textbf{72.6}  \\ 
			\bottomrule
		\end{tabular}%
	}
	\label{table7}%
\end{table}%

\begin{table*}[htbp]
	\scriptsize
	\setlength{\abovecaptionskip}{0cm}  %段前
	\setlength{\belowcaptionskip}{-0.0cm} %段后
	\centering
	\caption{Comparison with different ${\beta }^{t}$ on Office-Home with BiPC-R50}
	\renewcommand{\arraystretch}{1.2}
	\setlength{\tabcolsep}{0.5mm}{
		\begin{tabular}{c|ccccccccccccc}
			\toprule
			${\beta }^{t}$ (Source→Target)& {Ar→Cl} & {Ar→Pr} & {Ar→Re} & {Cl→Ar} & {Cl→Pr} & {Cl→Re} & {Pr→Ar} & {Pr→Cl} & {Pr→Re} & {Re→Ar} & {Re→Cl} & {Re→Pr} & \cellcolor{gray!25}{Avg.} \\
			\midrule
			$0.5$ & 58.4& 79.0 & 82.3 &68.4 & 78.3 & 77.9 & 68.0 & 55.3 & 80.3 & 72.5 & 60.0 & 84.1 & \cellcolor{gray!25}72.0\\
			$\exp \left(-\mathrm{GE}\left(p_{h}^{t}\right)\right)$ & 58.4 & 78.1 & 82.2 & 67.8 & 77.9 & \textbf{79.0} & 67.1 & 56.8 & 80.8 & 73.0& 60.1 & \textbf{85.1}&\cellcolor{gray!25}72.2\\
			$\mathrm{max}(p_{h}^{t})$ 	& 58.0 & 79.2 & \textbf{82.5} & \textbf{68.5} & \textbf{78.6} & \textbf{79.0} & \textbf{68.4} &56.8 & 81.0 & 72.6 & \textbf{61.4} &83.2 & \cellcolor{gray!25}72.4\\
			$\exp \left(-\mathrm{KL}\left(\tilde{p}_{h}^{t} \| p_{h}^{t}\right)\right)$	&\textbf{59.1}	&\textbf{79.3}	&82.2	&68.3	&78.4	&\textbf{79.0}	&67.3	&\textbf{57.1}	&\textbf{81.7}	&\textbf{74.0}	&60.3	&84.8	&\cellcolor{gray!25}\textbf{72.6}\\
			\bottomrule
		\end{tabular}%
	}
	\label{table8}%
\end{table*}% 

\begin{table*}[htbp]
	\scriptsize
	\setlength{\abovecaptionskip}{0cm}  %段前
	\setlength{\belowcaptionskip}{-0.0cm} %段后
	\centering
	\caption{Comparison with different penalties on Office-Home with BiPC-R50.}
	\renewcommand{\arraystretch}{1.1}
	\setlength{\tabcolsep}{0.5mm}{
		\begin{tabular}{c|ccccccccccccc}
			\toprule
			Method (Source→Target)& {Ar→Cl} & {Ar→Pr} & {Ar→Re} & {Cl→Ar} & {Cl→Pr} & {Cl→Re} & {Pr→Ar} & {Pr→Cl} & {Pr→Re} & {Re→Ar} & {Re→Cl} & {Re→Pr} & \cellcolor{gray!25}{Avg.} \\
			\midrule
			GE	& 56.1& 76.0 &79.1 &65.8 & 74.1 &73.4 & 66.7 & 54.8 & 79.8 & 72.3& 58.6 & 83.3 & \cellcolor{gray!25}70.0\\
			CGE	& 55.8 & 75.2 & 78.7 & 66.9 & 74.9 & 74.5 &66.8 & 55.3 & 79.0 & 73.4 & 58.6& 83.5& \cellcolor{gray!25}70.2\\
			GI	& 58.4& 78.5 & \textbf{82.3} &67.7 &78.7 & \textbf{79.4} &67.0 &56.3 & \textbf{82.1} & 72.8 & 59.4 &84.8 & \cellcolor{gray!25}72.3\\
			CGI (w/o reg)	&58.1 & 78.4 &81.3 &\textbf{68.4} & \textbf{79.1} &77.4 & \textbf{67.3} & \textbf{57.5} & 81.1 & 73.8 & 60.0 &\textbf{84.9} & \cellcolor{gray!25}72.3\\
			CGI	&\textbf{59.1}	&\textbf{79.3}	&82.2	&68.3	&78.4	&79.0	&\textbf{67.3}	&57.1	&81.7	&\textbf{74.0}	&\textbf{60.3}	&84.8	&\cellcolor{gray!25}\textbf{72.6}\\
			\bottomrule
		\end{tabular}%
	}
	\label{table9}%
\end{table*}% 

\subsection{Ablation Study}
\textbf{Contribution of the components}. BiPC comprises three components: standard classification, CPA, and CGI. To evaluate the individual contributions of each component to the overall approach, we conducted an additional experiment to assess the effectiveness of average accuracy on the Office-Home dataset using BiPC-R50. While the classification of the source domain is crucial, we mainly focus on investigating the impact of the proposed CPA and CGI. Notably, the feature extractor $\theta$ has not been trained in CGI. Therefore, in this section, we also examine the influence of whether $\theta$ is trained or not. We denote the feature extractor trained with CGI as CGI (${{\theta }_{h}}$ $\theta$) and the original CGI as CGI (${{\theta }_{h}}$). Table 7 clearly indicates that both CPA and CGI significantly contribute to the performance improvement. Additionally, we observe that training the feature extractor in CGI without CPA leads to a greater improvement. However, when combined with CPA, training the task head alone produces better results.

\textbf{Probability Space vs Feature Space}. Table 7 demonstrates that solely training the CPA significantly enhances model performance. This finding indicates that probability alignment not only furnishes more informative data for task-head training but also enhances the model representation. To the best of our knowledge, we are the first to consider matching source and target domain distributions in probability space, therefore, the comparison with the feature space is necessary to highlight the advantages of the probability space. Following the settings of (\cite{ref4}, \cite{ref6}), feature space is defined as the output of feature extractor, of which source and target defined as ${{f}^{s}}={{F}_{\theta }}\left( {{x}^{s}} \right)$ and ${{f}^{t}}={{F}_{\theta }}\left( {{x}^{t}} \right)$ respectively. A minor difference is that BottleNeck (\cite{ref4}, \cite{ref6}) is not used. Besides, the feature space alignment can be formulated as $\sum\nolimits_{i=1}^{{{n}_{s}}}{\sum\nolimits_{j=1}^{{{n}_{t}}}{\alpha _{ij}^{st} d\left( f_{i}^{s},f_{j}^{t} \right)}}$, as well as  $d$ is consider as a linear and Radial Basis Function (RBF). Taking the ``Art→Clipart'' task in the Office-Home dataset as an example, we separately analyze the ResNet-50 and DeiT-Base models while setting $\lambda_{2}$ to 0.1 due to the feature space's sensitivity. For simplicity, we do not consider CGI during the experiments. Figure 3 illustrates that the probability space exhibits significant advantages in terms of convergence and performance compared to both the feature space and the baseline. This phenomenon is observed in both CNN and Transformer structures. Hence, further exploration of probability space based UDA approaches is warranted in the future.

\textbf{Selection of ${{\beta}^{t}}$}. Given that the calibration factor of CGI is calculated through an adaptive strategy, the design of ${\beta}^{t}$ holds significant importance in influencing the model's performance. Since ${{\beta}^{t}}$ requires different responses to input instances and must be constrained within the range of 0 to 1, we have explored four common design ideas, which are outlined below: (a) $0.5$, (b) $\exp \left(-\mathrm{GE}\left(p_{h}^{t}\right)\right)$, (c) $\mathrm{max}(p_{h}^{t})$, and (d) $\exp \left(-\mathrm{KL}\left(\tilde{p}_{h}^{t} \| p_{h}^{t}\right)\right)$. CPA is maintained throughout the experiment, while all hyper-parameters are kept consistent except for ${{\beta}^{t}}$. Table 8 clearly demonstrates that incorporating the transformed probability $\tilde{p}_{h}^{t}$ into the calibration factor calculation yields the best performance, whereas the absence of an adaptive factor leads to poor results. Consequently, we have ultimately selected (d) $\exp \left(-\mathrm{KL}\left(\tilde{p}_{h}^{t} \| p_{h}^{t}\right)\right)$ as the calibration factor.

\textbf{CGI vs Different Penalization}. Unlike previous approaches, we utilize Gini impurity as the foundation and introduce CGI for training on unlabeled target data. In this section, we will discuss (a) Gibbs entropy (GE), (b) Gini impurity (GI), (c) calibrated Gibbs entropy (CGE), and (d) CGI. Additionally, we will analyze CGI without the regularization term $\left( 1-{{\beta }^{t}} \right) \text{GI}\left( p_{m}^{t} \right)$ in Eq. (16), denoted as (e) CGI (w/o reg). CPA is employed in all experiments as it involves the calculation of correction factors. Table 9 provides illuminating insights based on the experimental findings: (1) For pseudo-label learning, Gini impurity outperforms Gibbs entropy. (2) The incorporation of correction factors helps mitigate overconfidence in pseudo-labeling and enhances the performance across various penalty terms. (3) Regularization terms further improve the effectiveness of CGI.

\textbf{Further Networks Architectures}. Since the introduction of Transformers, there has been an enriched variety of network structures. To further investigate the flexibility and effectiveness of our approach, we conducted experiments on the VisDA-2017 dataset using a wider range of network structures, including ConvNeXt, CrossViT, gMLP. Table 10 demonstrates significant performance variations of different backbones on UDA tasks. For instance, the baseline of SwinT outperforms gMLP with BiPC. Moreover, BiPC exhibits further performance improvements across various backbones, showcasing the flexibility and effectiveness of our method.

% Table generated by Excel2LaTeX from sheet 'Table8'
\begin{table*}[htbp]
	\scriptsize
	\setlength{\abovecaptionskip}{0cm}  %段前
	\setlength{\belowcaptionskip}{-0.0cm} %段后
	\centering
	\caption{BiPC with further backbones based on VisDA-2017 for vanilla closed-set UDA.}
	\renewcommand{\arraystretch}{1.1}
	\begin{tabular}{c|ccccccccccccc}
		\toprule
		Method (Synthesis→Real)& plane & {bcycl} & {bus} & {car} & {horse} & {knife} & {mcycl} & {person} & {plant} & {sktbrd} & {train} & {truck} & \cellcolor{gray!25}{Avg.} \\
		\midrule
		Baseline-CB (\cite{ref68}) & \textbf{98.5} & 75.4  & \textbf{86.2} & 65.4  & 89.4  & 57.5  & \textbf{94.5} & 17.3  & 92.1  & 57.9  & \textbf{93.0} & 39.0  & \cellcolor{gray!25}72.2  \\
		Ours-CB & 98.3  & \textbf{86.0} & 84.6  & \textbf{69.2} & \textbf{95.8} & \textbf{97.7} & 93.8  & \textbf{69.4} & \textbf{95.8} & \textbf{85.7} & 92.6  & \textbf{64.2} & \cellcolor{gray!25}\textbf{86.1} \\
		\midrule
		Baseline-CVB (\cite{ref71}) & \textbf{97.6} & 64.5  & 76.7  & 62.7  & 75.5  & 68.1  & \textbf{93.6} & 11.9  & 73.4  & 27.0  & \textbf{92.8} & 20.2  & \cellcolor{gray!25}63.7  \\
		Ours-CVB & 97.5  & \textbf{86.0} & \textbf{84.6} & \textbf{63.2} & \textbf{96.2} & \textbf{95.7} & 92.7  & \textbf{77.3} & \textbf{92.8} & \textbf{73.4} & 91.0  & \textbf{60.8} & \cellcolor{gray!25}\textbf{84.3} \\
		\midrule
		Baseline-gMS (\cite{ref20}) & 95.9  & 70.8  & 62.8  & 75.4  & 90.9  & 39.5  & 90.4  & 36.3  & 84.6  & \textbf{11.4} & \textbf{91.3} & 24.0  & \cellcolor{gray!25}64.5  \\
		Ours- gMS & \textbf{96.8} & \textbf{82.0} & \textbf{79.9} & \textbf{81.3} & \textbf{94.6} & \textbf{85.9} & \textbf{92.6} & \textbf{67.9} & \textbf{94.4} & 2.50   & 88.9  & \textbf{53.2} & \cellcolor{gray!25}\textbf{76.6} \\
		\bottomrule
	\end{tabular}%
	\label{table10}%
\end{table*}%

% Table1
\begin{table*}[htbp]
	\scriptsize
	\setlength{\abovecaptionskip}{0cm}  %段前
	\setlength{\belowcaptionskip}{-0.0cm} %段后
	\centering
	\caption{Comparison with SoTA methods on Office-Home (65→25) for partial-set UDA.}
	\renewcommand{\arraystretch}{1.1}
	\setlength{\tabcolsep}{0.5mm}{
		\begin{tabular}{c|ccccccccccccc}
			\toprule
			Method (Source→Target)& {Ar→Cl} & {Ar→Pr} & {Ar→Re} & {Cl→Ar} & {Cl→Pr} & {Cl→Re} & {Pr→Ar} & {Pr→Cl} & {Pr→Re} & {Re→Ar} & {Re→Cl} & {Re→Pr} & \cellcolor{gray!25}{Avg.} \\
			\midrule
			\hspace{-0.7cm}\makecell[l]{\textit{ResNet50:}} &    &    &     &   &       &       &       &       &       &       &       &       & \cellcolor{gray!25} \\
			DRCN (\cite{ref74}) &54.0 &76.4 &83.0 &62.1 &64.5 &71.0 &70.8 &49.8 &80.5 &77.5 &59.1 &79.9 &\cellcolor{gray!25}69.0 \\
			BA$^{\text{3}}$US (\cite{ref75}) &60.6 &83.2 &88.4 &71.8 &72.8 &83.4 &75.5 &61.6 &86.5 &79.3 &62.8 &86.1 &\cellcolor{gray!25}76.0\\
			TSCDA (\cite{ref76}) &63.6 &82.5 &89.6 &73.7 &73.9 &81.4 &75.4 &61.6 &87.9 &83.6 &67.2 &88.8 &\cellcolor{gray!25}77.4\\
			SHOT (\cite{ref58})	 &64.6 &85.1 &\textbf{92.9} &78.4 &76.8 &\textbf{86.9} &79.0 &65.7 &\textbf{89.0} &81.1 &\textbf{67.7} &86.4 &\cellcolor{gray!25}79.5 \\
			Baseline-R50	& 50.5 & 73.0 & 82.4 & 63.4 & 68.2 & 75.4 & 57.9 &45.0 & 76.1 & 68.5  & 50.7 & 78.9 &\cellcolor{gray!25}65.8\\
			Ours-R50	&\textbf{66.3} 	&\textbf{86.3} 	&90.5 	&\textbf{78.8} 	&\textbf{79.8} 	&84.7 	&\textbf{82.5} 	&\textbf{67.4} 	&84.0 	&\textbf{85.4} 	&65.4 	&\textbf{88.7} 	&\cellcolor{gray!25}\textbf{80.0} \\
			\midrule
			\hspace{-0.7cm}\makecell[l]{\textit{ViT:}} &       &       &       &       &       &       &       &       &       &       &       &       &  \cellcolor{gray!25}\\
			Baseline-DB	& 61.6 & 80.8  & 88.3 & 74.9 & 72.7 & 80.0 &74.7 & 55.2 & 84.4 & 80.1 & 59.0 & 84.5 & \cellcolor{gray!25}74.7\\
			Ours-DB	& \textbf{73.6} & \textbf{88.5}  & \textbf{92.4} & \textbf{84.7} & \textbf{86.4} & \textbf{88.3} &\textbf{86.2} & \textbf{68.7} & \textbf{89.3} & \textbf{86.4} & \textbf{67.8} & \textbf{90.6} & \cellcolor{gray!25}\textbf{83.6}\\
			\midrule
			\hspace{-0.7cm}\makecell[l]{\textit{SwinT:}} &       &       &       &       &       &       &       &       &       &       &       &       &  \cellcolor{gray!25}\\
			Baseline-SB$^o$	& 76.2 & 85.9 & 92.4 & 85.4 & 85.8 & 88.6 & 84.6 &70.9 & 90.2 & 86.6 & 71.8 & 88.2 & \cellcolor{gray!25}83.9\\
			Ours-SB$^o$	& \textbf{81.2} & \textbf{89.0} & \textbf{95.1} & \textbf{90.5} & \textbf{89.4} & \textbf{92.8}& \textbf{90.1} & \textbf{81.3} & \textbf{94.5} & \textbf{91.4} & \textbf{77.4} & \textbf{92.3} & \cellcolor{gray!25}\textbf{88.8}\\
			\bottomrule
		\end{tabular}%
	}
	\label{table11}%
\end{table*}%

\subsection{Discussion}
\textbf{Convergence}. The experiments are conducted on the ``Art→Clipart'' task in the Office-Home dataset. We apply BiPC to DeiT-Base for comparison with CDTrans (\cite{ref14}), and BiPC to ResNet-50 to compare with ATDOC-NA (\cite{ref59}). To ensure fair comparison and highlight the fast convergence of our method, we follow the settings of (\cite{ref14}) and (\cite{ref59}). One epoch is defined as traversing the dataset once, and we train for 20 epochs, totaling 2420 iterations. In contrast, (\cite{ref14}) and (\cite{ref59}) are trained for 50 epochs, totaling 6050 iterations. Given the different training losses among methods, comparing based on test error is challenging. Hence, we use target domain classification accuracy as the basis for comparison. From Figure 4, although our method exhibits a slight performance degradation compared to Table 2, it clearly demonstrates that BiPC achieves significantly faster convergence than state-of-the-art methods.

\textbf{Partial-set Domain Adaptation}. In real-world scenarios, it is common for the categories in the target domain to be a subset of the source domain, which can be regarded as a special case of the traditional UDA, called Partial-set Domain Adaptation (PDA) (\cite{ref73}). To further validate the validity and flexibility of our method, we try to apply BiPC to the PDA task while only making simple modifications to the original method. When utilizing BiPC for PDA, the key aspect is to estimate the categories for the target dataset. This prevents the model from adapting to classes that are not part of the target domain, thereby avoiding potential performance degradation. Inspired by SHOT (\cite{ref58}), we count the number of predicted categories and estimate the categories on the target domain using a thresholding method. The counting of category $i$ can be formulated as follows:
\begin{eqnarray}
	n_{i}=\sum_{j=1}^{n_{t}} \mathbbm{1}_{\left[\text{argmax} \left(p_{h j}^{t}\right)=i\right]}.
\end{eqnarray}
The set of categories counting can be denoted as: ${{N}_{c}}=\{n_{i}\}_{i=1}^{{{c}_{1}}}$. Next, PDA can be implemented with BiPC by making a simple modification to the predicted probability of the samples $x_{j}^t$, represented as follows:
\begin{eqnarray}
	\hat{p}_{h j c}^{t}= \mathbbm{1}_{\left[n_{c}\ge T_{pda} \right]}p_{h j c}^{t}.
\end{eqnarray}
where $\hat{p}_{h j c}^{t}$ represents the modified probability of category $c$, $p_{h j c}^{t}$ denotes the original probability of category $c$, $n_{c}$ signifies the counting of categories $c$, and $T_{pda}$ is a predefined threshold. The PDA task can be realized by employing BiPC with the modified probability instead of the original probability. During the experiment, to attain improved performance, we set ${{\lambda }_{1}}$ to 0.5 and $a$ of ${{\lambda }_{3}}$ to 0.025. The impact of the choice of $T_{pda}$ on the performance is illustrated in Figure 5, and ultimately, $T_{pda}$ is selected to be 14.

For the experiment settings, we adopt the protocol outlined in (\cite{ref74}) for the Office-Home dataset. Specifically, the target domain in Office-Home consists of a total of 65 classes, and we focus on 25 classes (the first 25 in alphabetical order) for our analysis. BiPC is compared with the SoTA method for PDA, respectively:	DRCN (\cite{ref74}), BA$^{\text{3}}$US (\cite{ref75}), TSCDA (\cite{ref76}), SHOT (\cite{ref58}). As observed from Table 11, it is evident that BiPC demonstrates exceptional performance in the PDA task, achieving state-of-the-art results. Moreover, this highlights the versatility of BiPC in the PDA task, as it remains effective across different architectures.

\begin{figure}[t]
	\setlength{\abovecaptionskip}{0.cm}
	\setlength{\belowcaptionskip}{-0.cm}
	\centering
	\includegraphics[width=3.3in]{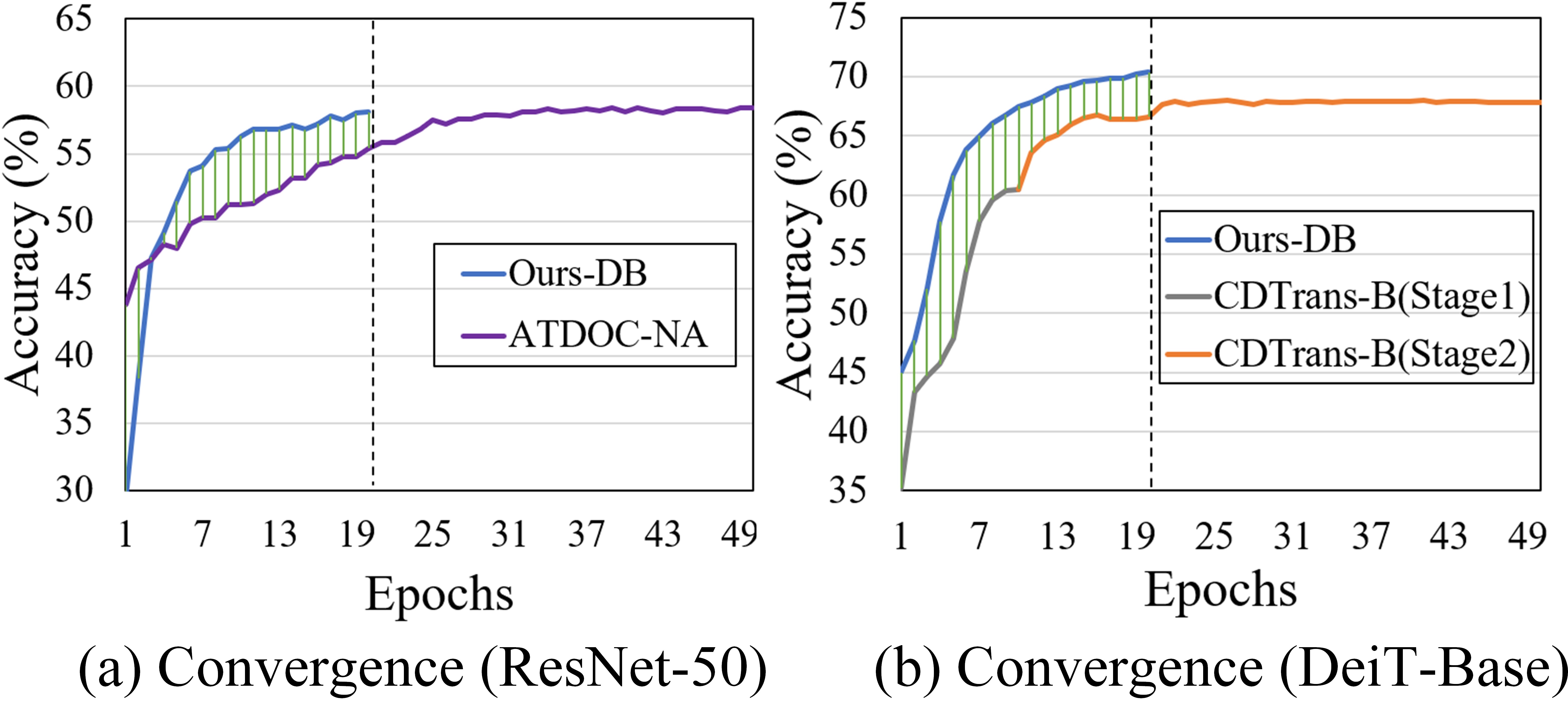}
	\caption{On task Art→Clipart (Office-Home), we further analyze the convergence of BiPC compared with SoTA. (a) and (b) are the convergence about ResNet-50 and DeiT-Base relevant methods respectively.}
	\label{fig4}
\end{figure}

\begin{figure}[t]
	\setlength{\abovecaptionskip}{0.cm}
	\setlength{\belowcaptionskip}{-0.cm}
	\centering
	\includegraphics[width=2.0in]{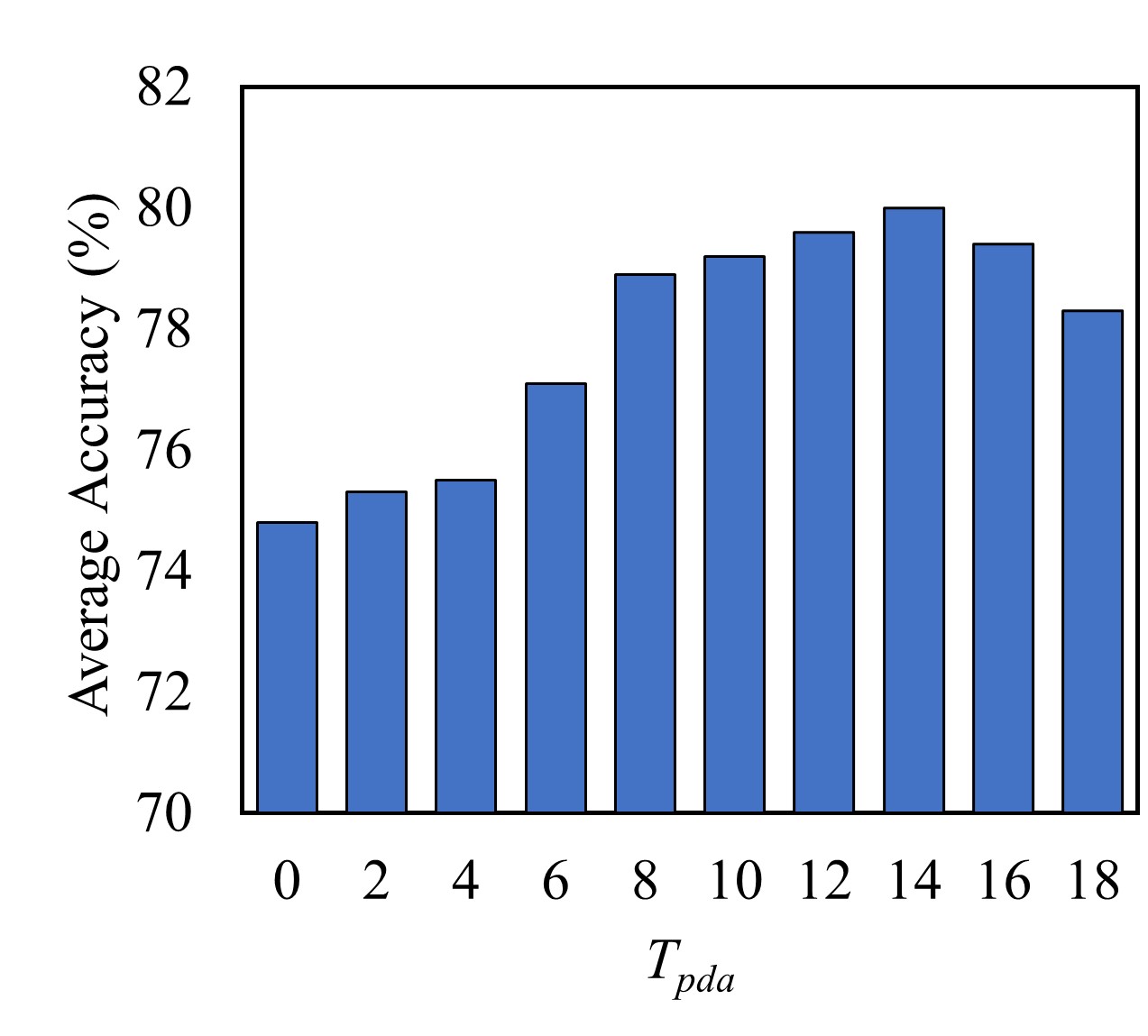}
	\caption{Average accuracy (\%) of BiPC for all partial-set DA tasks on Office-Home (ResNet-50).}
	\label{fig5}
\end{figure}

\section{Conclusion}
In this paper, we propose Bidirectional Probability Calibration (BiPC) to address the limited improvement of feature alignment and the lack of flexibility in Transformer-based methods. BiPC offers a simple yet effective approach that can be seamlessly integrated with popular backbones, benefiting both Transformer and CNN architectures in unsupervised domain adaptation. Through extensive experiments and various ablation studies, we demonstrate that BiPC achieves state-of-the-art or comparable performance on different datasets.

Despite its advantages, our work currently considers only supervised pre-training models, limiting the broader applicability of BiPC, especially with the increasing popularity of weakly supervised pre-training (e.g., CLIP (\cite{ref83})) and self-supervised pre-training (e.g., MAE (\cite{ref82})). Future research will focus on extending the generality of our approach to cover a wider range of models and task settings, including those leveraging weakly and self-supervised pre-training paradigms. Additionally, the current implementation of BiPC may require further refinement to ensure optimal integration with various backbone architectures and to address potential scalability issues when applied to large-scale datasets.

BiPC not only exhibits effectiveness and flexibility but also highlights the immense potential of the probability space in unsupervised domain adaptation. This study reveals that, compared to the commonly focused feature space, the probability space provided by pre-trained task heads demonstrates greater robustness to domain gaps. We believe this insight is valuable for researchers in this field, suggesting that future work could explore designing more effective UDA methods within the probability space, thereby better utilizing pre-trained task heads to enhance performance. Furthermore, the idea of leveraging additional supervisory information from pre-trained task heads could be extended to other domains, such as supervised learning and semi-supervised learning. It also holds promise for other visual tasks, including object detection and segmentation. Exploring these directions could lead to significant advancements and novel methodologies in the broader field of machine learning and computer vision.

We anticipate that the probability space of pre-trained models will be further explored by future researchers, potentially uncovering new methodologies and applications that can leverage this robust and flexible approach for improved performance across a variety of tasks and domains.

\section{Declaration of competing interest}
The authors declare that they have no known competing financial interests or personal relationships that could have appeared to influence the work reported in this paper.

\section{Acknowledgment}
\begin{sloppypar}
	The work is supported by the National Key Research and Development Program of China (2022YFF0607001), Guangdong Basic and Applied Basic Research Foundation (2023A1515010993), Guangdong Provincial Key Laboratory of Human Digital Twin (2022B1212010004), Guangzhou City Science and Technology Research Projects (2023B01J0011), Jiangmen Science and Technology Research Projects (2021080200070009151), Shaoguan Science and Technology Research Project (230316116276286), Foshan Science and Technology Research Project (2220001018608), Zhuhai Science and Technology Research Project (2320004002668).
\end{sloppypar}


\bibliographystyle{cas-model2-names}
%\bibliographystyle{unsrt}
\bibliography{reference.bib}

% % Biography
% % Here goes the biography details.

% % Here goes the biography details.

\end{document}